\documentclass[preprint,journal]{vgtc}       

\ifpdf
  \pdfoutput=1\relax                   
  \usepackage{graphicx}                
  \DeclareGraphicsExtensions{.pdf,.png,.jpg,.jpeg} 
\else
  \usepackage{graphicx}                
  \DeclareGraphicsExtensions{.eps}     
\fi%

\graphicspath{{figures/}{pictures/}{images/}{./}} 

\usepackage{microtype}                 
\PassOptionsToPackage{warn}{textcomp}  
\usepackage{textcomp}                  
\usepackage{mathptmx}                  
\usepackage{times}                     
\usepackage{cite}                      
\usepackage{tabu}                      
\usepackage{booktabs}                  

\usepackage[utf8]{inputenc} 
\usepackage[T1]{fontenc}    
\usepackage[colorlinks=true]{hyperref}       
\usepackage{url}            
\usepackage{amsfonts}       
\usepackage{nicefrac}       
\usepackage{soul}
\usepackage{color}
\usepackage{subfig}
\usepackage{graphicx}
\usepackage{diagbox}
\usepackage[table]{xcolor}
\usepackage{comment}
\usepackage{multirow}
\usepackage{mathtools}
\usepackage{wrapfig}

\definecolor{mathias_color}{rgb}{.6,.4,.05}
\definecolor{edited_color}{rgb}{.5,.7,.1}
\definecolor{chengcheng_color}{rgb}{0.35,0,0}
\definecolor{chris_color}{rgb}{0,0.35,0}
\definecolor{yuanlu_color}{rgb}{0,0,0.35}
\definecolor{yijing_color}{rgb}{0,0.35,0.35}
\definecolor{rob_color}{rgb}{0.35,0.35,0}
\definecolor{lingling_color}{rgb}{0.35,0,0.35}
\definecolor{todo_color}{rgb}{1.0,0,0.0}

\newcommand{\dataset}{UNOC\xspace}
\DeclarePairedDelimiterX{\norm}[1]{\lVert}{\rVert}{#1}

\newcommand{\beforefigcaption}{}
\newcommand{\afterfigcaption}{}
\newcommand{\beforetab}{}
\newcommand{\aftertab}{}

\newcommand{\beforesubsection}{}
\newcommand{\aftersubsection}{}

\usepackage{xspace}
\newcommand*{\eg}{\textit{e.g.}\@\xspace}
\newcommand*{\ie}{\textit{i.e.}\@\xspace}
\newcommand*{\etal}{\textit{et al.}\@\xspace}
\makeatletter
\newcommand*{\etc}{%
    \@ifnextchar{.}%
        {etc}%
        {etc.\@\xspace}%
}
\makeatother

\usepackage{etoolbox}
\newcommand{\zerodisplayskips}{%
  \setlength{\abovedisplayskip}{3pt}%
  \setlength{\belowdisplayskip}{3pt}%
  \setlength{\abovedisplayshortskip}{0pt}%
  \setlength{\belowdisplayshortskip}{0pt}}
\appto{\normalsize}{\zerodisplayskips}
\appto{\small}{\zerodisplayskips}
\appto{\footnotesize}{\zerodisplayskips}

\makeatletter
\newcommand{\thickhline}{%
    \noalign {\ifnum 0=`}\fi \hrule height 1pt
    \futurelet \reserved@a \@xhline
}
\makeatother

\onlineid{1071}

\vgtccategory{Research}
\vgtcpapertype{algorithm/technique}

\title{UNOC: Understanding Occlusion for Embodied Presence in Virtual Reality}

\author{Mathias Parger~*, Chengcheng Tang~\textdagger, Yuanlu Xu~\textdagger, Christopher Twigg~\textdagger, \\ Lingling Tao~\textdagger, Yijing Li~\textdagger, Robert Wang~\textdagger, and Markus Steinberger~*}
\authorfooter{
\item *~Graz University of Technology.
\item \textdagger~Facebook Reality Labs
}

\shortauthortitle{Parger\MakeLowercase{\textit{et al.}}: UNOC: Understanding Occlusion for Embodied Presence in Virtual Reality}

\abstract{Tracking body and hand motions in the 3D space is essential for social and self-presence in augmented and virtual environments. Unlike the popular 3D pose estimation setting, the problem is often formulated as inside-out tracking based on embodied perception (\eg, egocentric cameras, handheld sensors).
In this paper, we propose a new data-driven framework for inside-out body tracking, targeting challenges of omnipresent occlusions in optimization-based methods (\eg, inverse kinematics solvers).
We first collect a large-scale motion capture dataset with both body and finger motions using optical markers and inertial sensors. This dataset focuses on  social scenarios and captures ground truth poses under self-occlusions and body-hand interactions.
We then simulate the occlusion patterns in head-mounted camera views on the captured ground truth using a ray casting algorithm and learn a deep neural network to infer the occluded body parts.
In the experiments, we show that our method is able to generate high-fidelity embodied poses by applying the proposed method on the task of real-time inside-out body tracking, finger motion synthesis, and 3-point inverse kinematics. %
} 


\CCScatlist{ 
}

\begin{document}



\maketitle

\section{Introduction}

Thanks to fast wireless Internet and ubiquitous smartphones, an ever-increasing fraction of communication takes place virtually.  This is even more true during social distancing measures in a pandemic, which have made video calling the {\em de facto} standard for everything from corporate meetings to Ph.D defenses.  However, among other limitations, video calling lacks the sense of being physically present in the same space.  The use of avatars in virtual reality (``embodied VR'') is one promising direction which simulates co-presence in a virtual 3D space through spatial audio and body language~\cite{SmithEmbodiedVR}.  

Because complicated capture setups requiring outside-in cameras are impractical, the industry is converging on self-contained headsets where cameras mounted on the headset (``inside-out'') can be used for both headset and controller tracking.  While recent progress has been made using these cameras to track hands~\cite{Bambach_2015_ICCV,MANO2017,OccludedHands17}, tracking the full body from the same cameras remains a challenge~\cite{EgoBodyTrack15,EgoCap16,xu2019mo2cap2}.  Headset-mounted cameras suffer from both occlusion and a limited field of view.  When standing, the upper body may occlude the lower body; when seated, the knees may occlude the feet.  Users may interlace their hands behind their heads or reach up to scratch their cheek.  

To track these challenging cases where visual evidence is lacking, we need strong priors to generate plausible motion from sparse data.  Building these priors requires high quality motion data.  Existing motion capture databases focus on dynamic motions found in gaming or visual effects scenarios: walking, running, acrobatic jumps.  Such motions are uncommon in VR meeting settings.  

\begin{figure*}[ptb]
\centering
\subfloat{\includegraphics[width=0.135\linewidth]{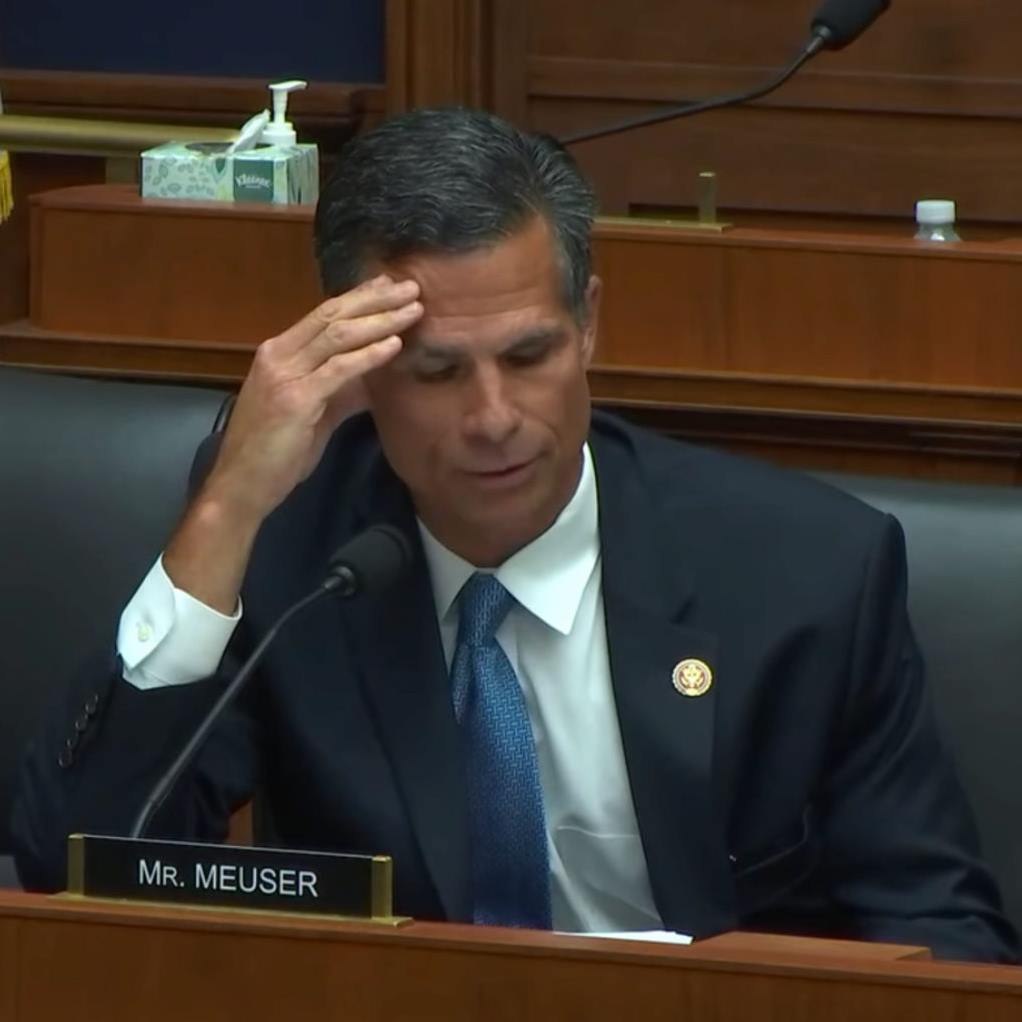}}
\subfloat{\includegraphics[width=0.135\linewidth]{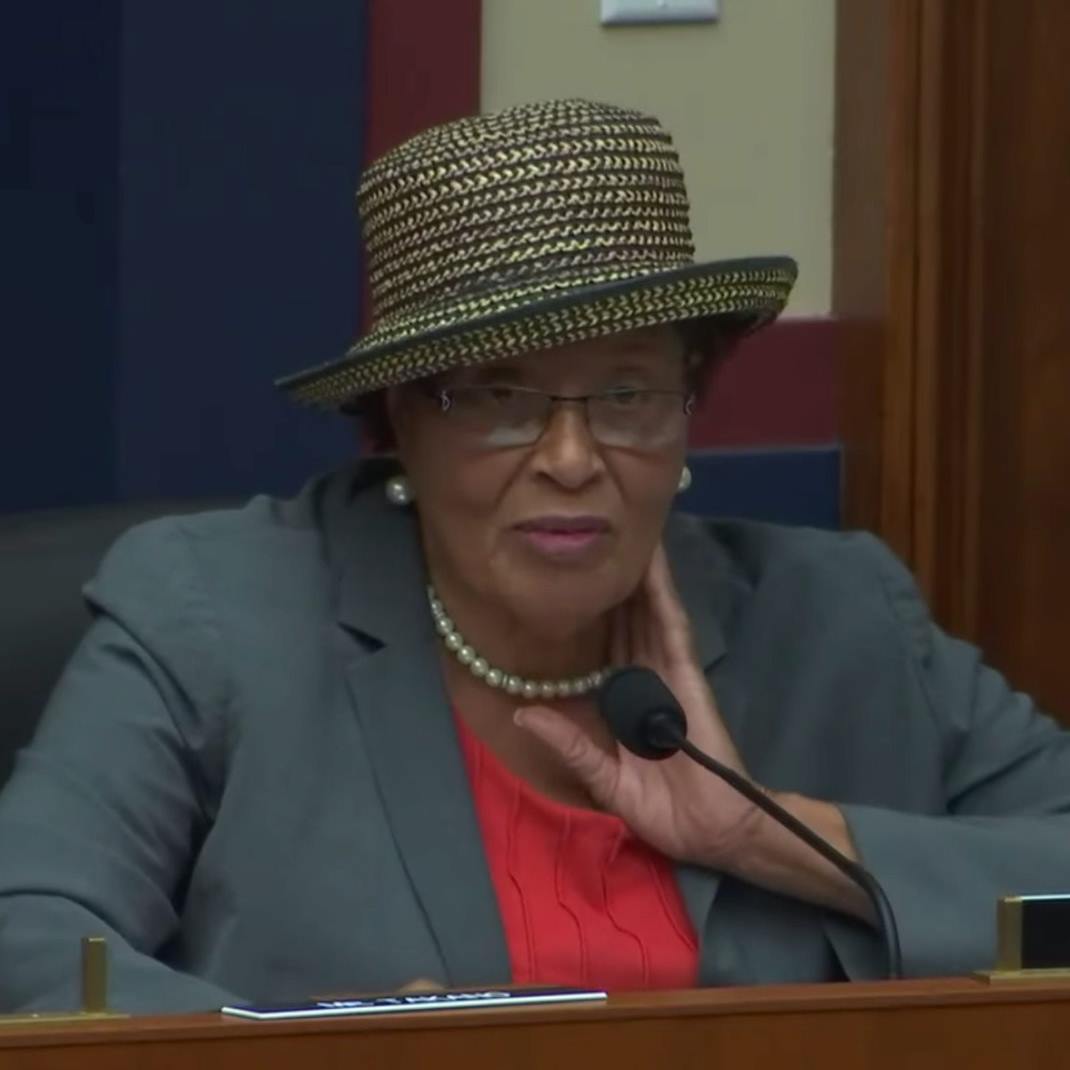}}
\subfloat{\includegraphics[width=0.135\linewidth]{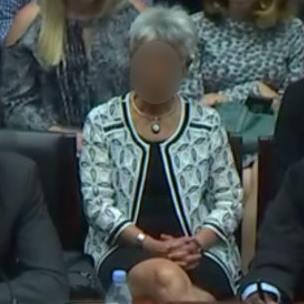}}
\subfloat{\includegraphics[width=0.135\linewidth]{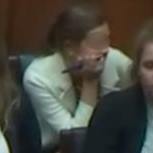}}
\subfloat{\includegraphics[width=0.135\linewidth]{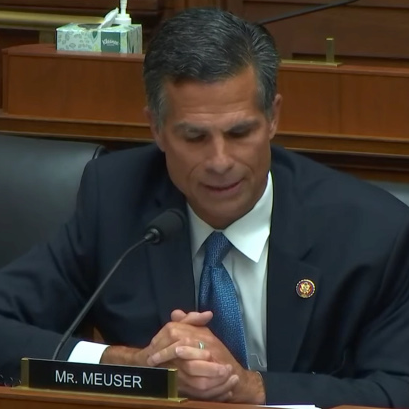}}
\subfloat{\includegraphics[width=0.135\linewidth]{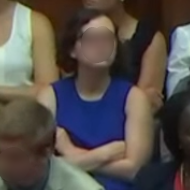}}
\subfloat{\includegraphics[width=0.135\linewidth]{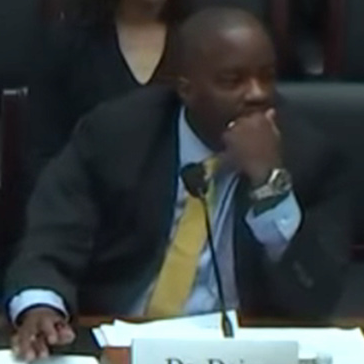}}
\vspace{-2mm}
\subfloat{\includegraphics[width=0.105\linewidth]{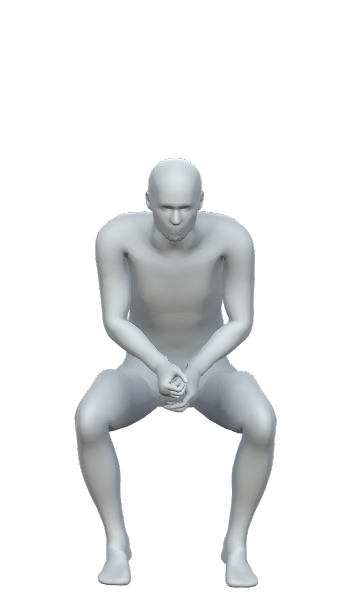}}
\subfloat{\includegraphics[width=0.105\linewidth]{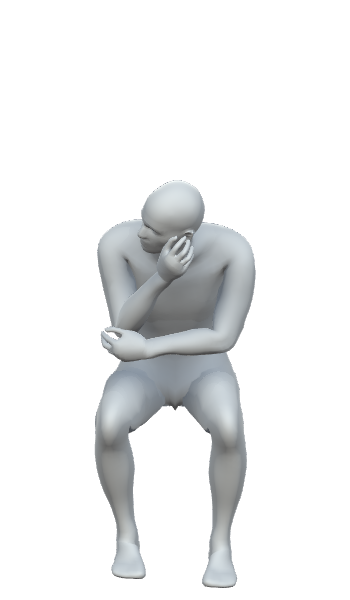}}
\subfloat{\includegraphics[width=0.105\linewidth]{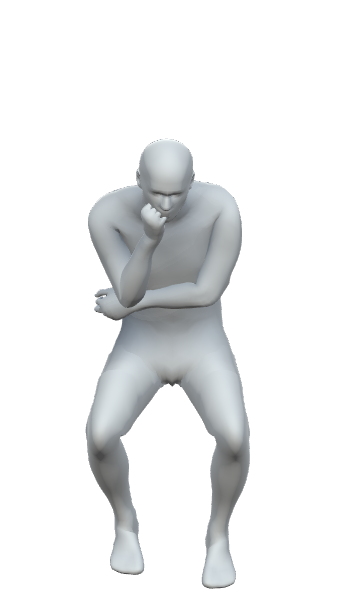}}
\subfloat{\includegraphics[width=0.105\linewidth]{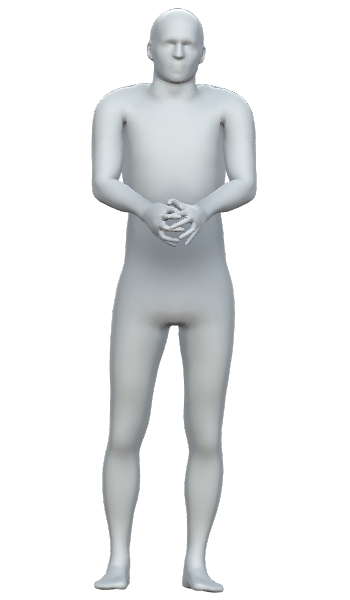}}
\subfloat{\includegraphics[width=0.105\linewidth]{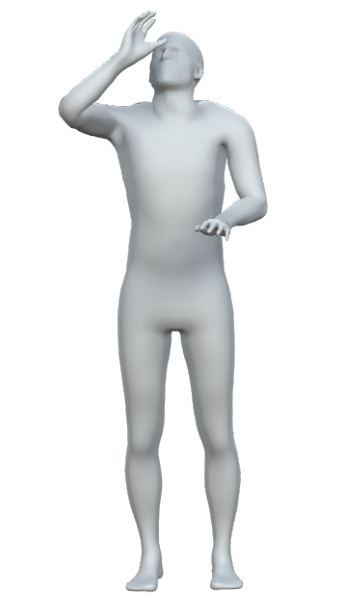}}
\subfloat{\includegraphics[width=0.105\linewidth]{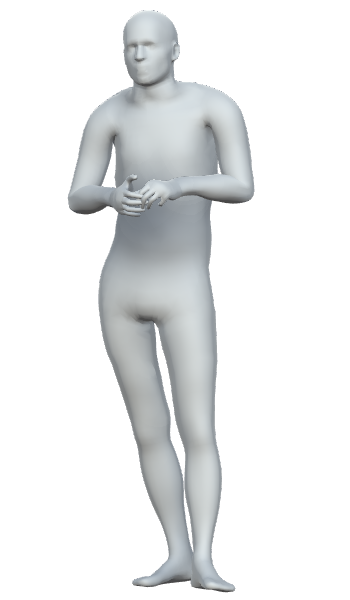}}
\subfloat{\includegraphics[width=0.105\linewidth]{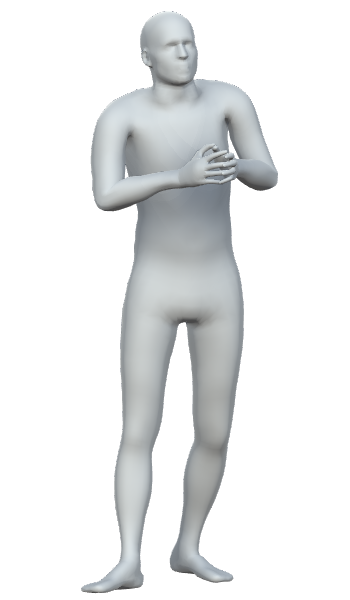}}
\subfloat{\includegraphics[width=0.105\linewidth]{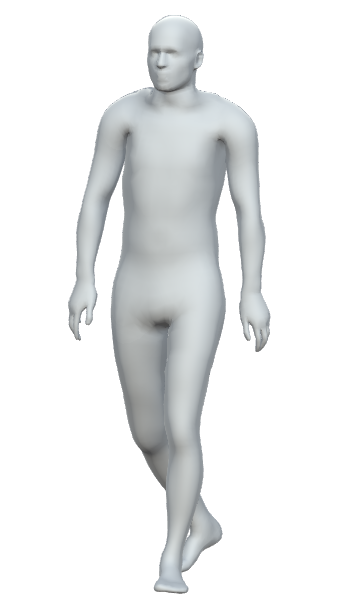}}
\subfloat{\includegraphics[width=0.105\linewidth]{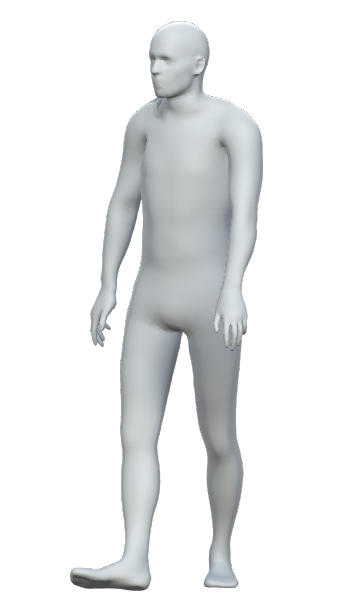}}
\beforefigcaption
\caption{Example poses from our study (bottom) and from our collected dataset of public domain CSPAN videos (top).  Notice the frequency of self-contact in the CSPAN data including clasped hands, crossed arms, and hand-face contact.  In our study, we collect poses ``inspired'' by 
the CSPAN data, resulting in a dataset of challenging cases that can be used to train and test machine learning models.}
\afterfigcaption
\label{fig:pose_example}
\end{figure*}

A survey of public domain CSPAN videos (see Fig.~\ref{fig:pose_example}) shows that meeting attendees display a variety of behaviors seldom captured on the traditional motion capture stage.  They may slouch, cross their arms, rest a cheek against a palm, scratch an ear or press their hands between their knees.  These motions have several characteristics in common.  First, they involve a combination of the hand and body: capturing only the joints above the wrist will mean missing out on important context.  Second, self-contact is an important part of the motion: the hand touches a cheek, or the hands are tucked inside armpits or pockets.  Third, much of the motion is occluded and potentially out of frustum for a typical headset camera configuration, making it particularly challenging for inside-out body trackers.

Capturing these motions is even difficult using traditional optical motion capture setups. Due to occlusions, markers are constantly disappearing, and entire limbs may be mis-predicted under a lack of enough markers. We address this challenge in several ways.  First, we use a non-line-of-sight IMU-based glove capture system for the hands.  Second, we send all captured data (including RGB videos) to a motion capture studio for manual clean-up.  Finally, we clean up the final motion with an inverse kinematics algorithm.  To adapt the dataset for inside-out body tracking, we propose a simulation algorithm to generate occlusion labels about what headset camera views can observe.


In addition to creating a high quality dataset of meeting motions, we propose a method to predict occluded or out-of-view limbs for inside-out body tracking.
Specifically, we apply temporal neural networks (\ie, RNN, LSTM~\cite{LSTM}, GRU~\cite{GRU}, TCN~\cite{pavllo2019}) to predict ground truth locations from poses observed in a time window.  Our method is designed to be a differentiable post-process for inside-out body trackers, useful as a pose prior that regularizes the output motion.



Evaluating our method on both existing public datasets and our newly collected motion, we show that we can generate plausible embodied poses for both the body and the hands with substantial missing information.  We also conduct detailed comparisons of the proposed model applied to two other popular tasks: 3-point inverse kinematics for VR, and finger motion synthesis, which shows that our method overcomes the problem of missing or occluded body joints during realistic interactions between the hands and the body.
Our experiments indicate that our data fills a gap in the existing collection of motion capture datasets and enables future research in learning embodiment for social presence in AR and VR.

\paragraph{Contributions} With the goal of predicting the pose of missing body joints, we make two major contributions:
\begin{enumerate}
    \item We collect a large-scale motion capture dataset (\dataset) for inside-out body tracking. The dataset focuses on difficult-to-track motions with hand-body interactions that are common in meetings and other social interactions.
    \item We propose a new deep neural network to predict the missing or occluded body parts and demonstrate that our model can generate plausible embodied poses in three applications: inside-out body tracking, finger motion synthesis, and 3-point inverse kinematics.
\end{enumerate}


\section{Related Work}


\paragraph{Body motion capture datasets}
Many body motion capture datasets are now available for research.
HumanEva~\cite{HumanEva10} and HDM05~\cite{Muller} contain simple human motions such as running, dancing and sports activities.  The CMU Motion Capture Dataset~\cite{Guerra-Filho2012} provides a larger variety of human motions including two-person interactions.
Human3.6M~\cite{Ionescu2014} includes more remote communication scenarios, such as face-to-face discussion and talking on the phone.
However, communication-focused poses are limited and all these dataset lack finger poses, which are important for VR presence~\cite{Wang2016Assessing}.
In contrast, our dataset focuses on contact-rich scenarios and provides both hand and body pose.

\paragraph{Body and hand motion capture datasets}
The paucity of motion data capturing both body and hands is best illustrated by AMASS~\cite{Mahmood}, which aggregates research motion datasets.
Out of the 15 included datasets spanning 40 hours, only the 8-minute-long \emph{TCD Hands}~\cite{Hoyet2012} provides the hand pose.
TCD Hands focuses on hand-object manipulation, pointing and signing using the hands, but contains little body-hand contact.
Recently, Taheri \etal~\cite{GRAB:2020} released \emph{GRAB}, a high-quality full-body optical motion capture dataset recording how humans hold different objects.  The resulting motion is designed to maximize the quality of the hand grasping and thus largely avoids the body self-occlusions seen in Fig.~\ref{fig:pose_example}.
Lee \etal~\cite{Lee2019} released a dataset incorporating body and hand tracking, \emph{TWH16.2M}. 
This dataset is a large-scale collection of unscripted conversations between multiple, simultaneously tracked participants, aiming to capture natural gestures happening during conversations.
However, the use of markers on the hands means that they cannot be captured when occluded by e.g. crossing the arms, and the resulting motion contains very few of the contact-rich motions that we focus on (see Table~\ref{tab:dataset_stats} for a comparison).
We capture a mix of scripted and unscripted motions with markerless gloves to achieve a diverse set of contact-rich poses as well as natural motion during conversations.

\paragraph{Occlusion-aware and temporal body tracking}
Explicitly modeling occlusion between body parts~\cite{sigal2006measure, 6751345} and people~\cite{ghiasi2014parsing} has been shown to improve the accuracy of human pose estimation.  
Various pose priors have also been used to address occlusions, including Mocap-guided priors~\cite{MoCapGuidePose_NIPS2016}, kinematic body models~\cite{kanazawa2018end}, adversarial poses discriminators~\cite{Chen2017Adversarial} and adversarial generation of occlusion masks during data augmentation~\cite{peng2018jointly}.  
Temporal information has also been shown to be effective at filling in missing information, and has been particularly successful at predicting depth from 2D joints~\cite{rayat2018exploiting,pavllo2019,NIPS2019_gberta,Sim2real19}.  VIBE~\cite{kocabas2019vibe} combines adversarial learning and temporal modeling by using a discriminator trained on AMASS~\cite{Mahmood} to distinguish between real human motions and the output of temporal pose and shape regression networks.
Cheng \etal~\cite{Cheng2019, Cheng20203DHP} combine both explicit occlusion modeling and temporal networks to filter out occluded 2D joint predictions before lifting them into 3D with temporal convolution networks.
We also propose a method of using explicit occlusion modeling and temporal neural networks, but with an application to inside-out body tracking.  Our method can serve as an additional pose prior to existing 3D human pose estimation techniques.

\paragraph{Egocentric body tracking} estimates body motion from egocentric cameras, which is both highly relevant for embodied presence and also presents highly challenging occlusion.
With the advancement of AR/VR technologies, inside-out body tracking using all-in-one headsets has received increasing attention in recent years. One line of work aims to learn plausible body pose without requirement of camera coverage of user's body in egocentric view. 
Yuan et al.~\cite{EgoPoseImitLearn18} propose to use imitation-learning to learn a video-conditioned control policy and demonstrate that the framework can predict plausible body pose without observing camera wearer's body. Jiang et al.~\cite{EgoVid3DPose17} leverage the static scene vs dynamic scene classifier, the dynamic motion signature of the surrounding scene as well as pose coupling in longer time span to form a global optimization problem to infer the pose sequence. Both frameworks assume invisible pose and rely on the assumption of correlation of the user's body motion with dynamics of the surrounding scene, and have been demonstrated only on simple body motions. 
When body parts are at least partially visible from the cameras, 3D body pose estimation can be applied.  
Xu \etal~\cite{xu2019mo2cap2} use a cap-mounted fish eye camera to capture the body and predict body joint heatmaps and depth, followed by back-projection into 3D camera space to obtain 3D body pose.
Tome \etal~\cite{xREgoPose19} also predict body joint heatmaps but adapt an encoder with a dual-branch decoder to transform joint heatmaps into 3D pose while accounting for uncertainty of 2D keypoint locations.
We consider our work to be orthogonal to and complement these vision-based methods by 1) providing hand and body motion datasets for synthesizing hard-to-capture data for challenging but critical body poses; and 2) as a motion prior that can be incorporated into these methods.


\paragraph{Finger synthesis and 3-point body tracking}
We evaluate our method on two additional missing data problems:  Finger synthesis predicts finger motion from body motion and has been addressed with database driven methods~\cite{Jorg2012,Mousas2015} and deep learning~\cite{Lee2019}.  3-point body tracking addresses the common case of VR headsets that track only the head and the two controllers held by the hands~\cite{Parger2018}.  We show that both our proposed dataset and motion prior are effective for these tasks.
\section{Proposed Dataset}
\paragraph{Dataset overview} With a focus on occlusion and contact during social interactions, we built a hand-body mocap pipeline that is robust to occlusion and recorded 220 minutes of motion with 13 adult participants (10 males and 3 females).
The dataset comprises 1.6M frames of cleaned data (plus 200K frames for T-poses and other calibration data) at 120 frames per second with 211 recorded sequences (besides 26 for calibration).
It captures frequent occlusion and heavy contact between hands, arms, head and other body parts during social interaction, sitting or standing, with or without a table.
We measure the prevalence of body-body part contacts in our data by slightly increasing the size of the primitives used for occlusion simulation (see \ref{sec:occlusion_simulation}) and detecting inter-penetrations between non-adjacent bones.
Compared to MIXAMO \cite{Xu2019DenseRaC} and TWH16.2M\cite{Lee2019}, \dataset has 2x and 10x more body-body part contacts.
The hands interact with each other and with the body in 59\% of the frames, compared to 24\% and 4\%.
Detailed statistics and comparisons are shown in Table~\ref{tab:dataset_stats}.


\beforesubsection
\subsection{Motion capture pipeline}
\begin{figure}[ptb]
\centering
\includegraphics[width=\linewidth]{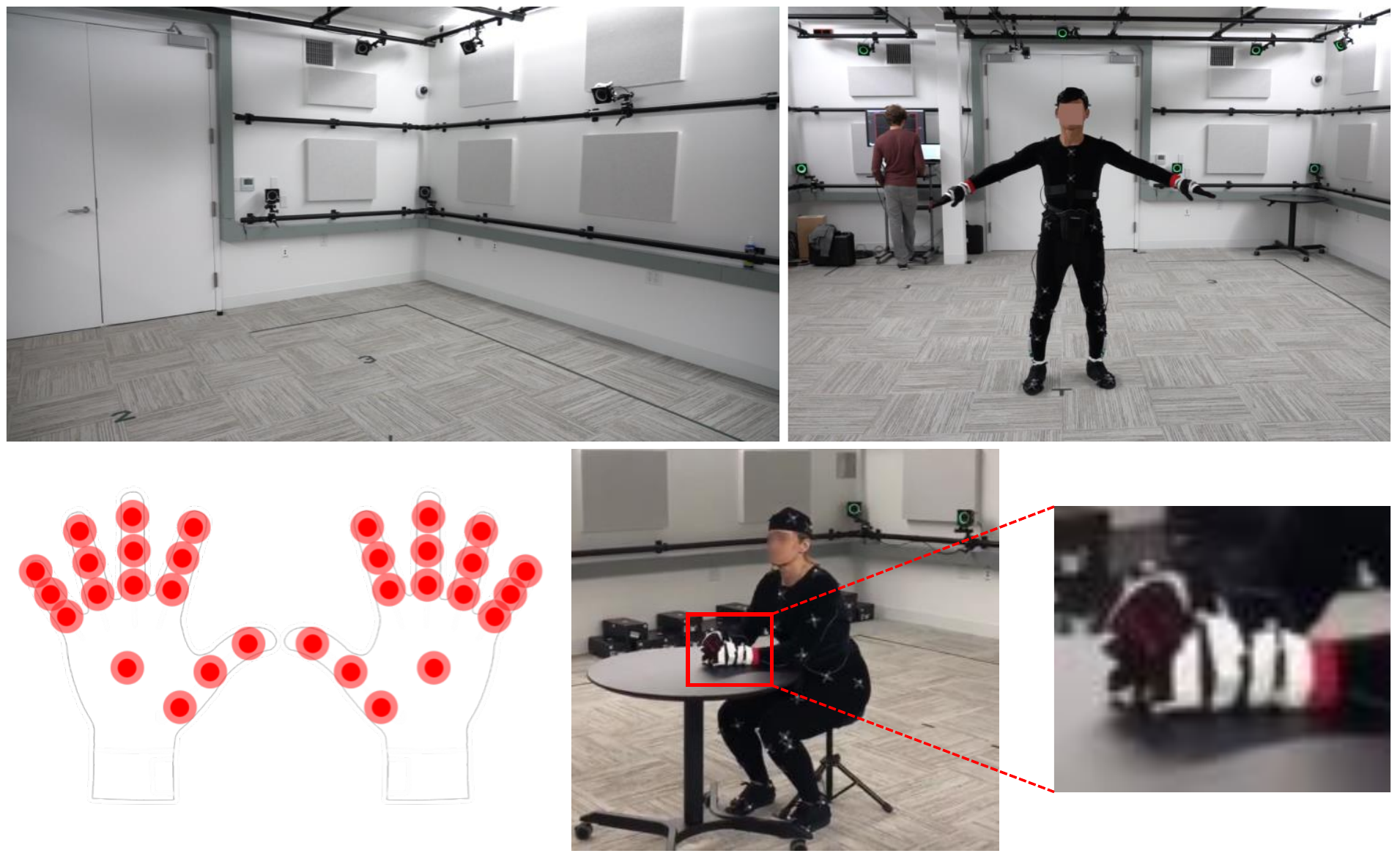}
\beforefigcaption
\caption{Our motion capture setup consists of Optitrack marker suit and IMU-based gloves. IMU-based devices show better properties in capturing challenging self-occluding and interacting poses.}
\afterfigcaption
\label{fig:mocap}
\end{figure}

\aftersubsection
To capture occlusion- and contact-heavy motion, we combine traditional optical motion capture (for the body) with an IMU-based system (for the hands).  As shown in Fig.~\ref{fig:mocap}, our optical tracking system uses 23 Optitrack Prime 22 cameras to capture 57 retro-reflective markers placed around the subject's body using the Optitrack Biomech marker configuration.
The tracked markers are manually cleaned and verified for all sequences to prevent mislabeling due to occlusion.
The used NANSENSE BioMed gloves utilize 16 IMU sensors for each hand to track the orientation of palm and finger segments. The tracked hand poses are robust to severe or complete occlusion, with acceptable compromises regarding latency and accuracy, and the lack of a calibrated global transform  with optical tracking.

We combine the solved body and hand motions by using the finger joint transforms from the glove-tracked hand, while using the hand orientation from the optical system.
We found that the optical-tracked hand orientation has a different yet consistent offset for each participant due to the placement of sparse markers on the wrists and hands.
We remove the differences by an additional calibration sequence in which we carefully place retro-reflective markers on the fingertips and compute the offset by rotating and aligning the optical-tracked hands with the auxiliary markers.


\beforesubsection
\subsection{Data capture protocol}
\aftersubsection

\begin{table*}[htb]
\centering
\setlength{\tabcolsep}{16pt}
\renewcommand\arraystretch{1.15}
\resizebox{\linewidth}{!}{
\begin{tabular}{c|c|c|c|c|c|c}
\hline\thickhline
Statistics & \multicolumn{2}{c|}{\dataset} & \multicolumn{2}{c|}{TWH16.2M~\cite{Lee2019}} & \multicolumn{2}{c}{Mixamo~\cite{Xu2019DenseRaC}}\\
\hline
Participants & \multicolumn{2}{c|}{13 (3 female)} & \multicolumn{2}{c|}{50} & \multicolumn{2}{c}{-}\\
\# Markers & \multicolumn{2}{c|}{57 + 32} & \multicolumn{2}{c|}{85} & \multicolumn{2}{c}{-}\\
FPS & \multicolumn{2}{c|}{120}  & \multicolumn{2}{c|}{90}    & \multicolumn{2}{c}{30}\\
\# Frames & \multicolumn{2}{c|}{1.8M} & \multicolumn{2}{c|}{16.2M} & \multicolumn{2}{c}{261K}\\
Capture Device & \multicolumn{2}{c|}{OptiTrack, Inertial} & \multicolumn{2}{c|}{OptiTrack} & \multicolumn{2}{c}{CGI Artists}\\
Captured Motion & \multicolumn{2}{c|}{Body, Hand} & \multicolumn{2}{c|}{Body, Hand} & \multicolumn{2}{c}{Body, Hand}\\
Content & \multicolumn{2}{c|}{Social Interaction} & \multicolumn{2}{c|}{Conversation} & \multicolumn{2}{c}{Dance, Action}\\
Self-contact & \multicolumn{2}{c|}{24\%} &  \multicolumn{2}{c|}{2.5\%}  &  \multicolumn{2}{c}{13\%}  \\
Hand-body-contact & \multicolumn{2}{c|}{59\%} &  \multicolumn{2}{c|}{3.9\%}  &  \multicolumn{2}{c}{24\%}  \\
\hline\thickhline
Occlusion & Ratio & Avg. Duration & Ratio & Avg. Duration & Ratio & Avg. Duration\\
\hline
Body        & 35\% & 3.02s & 38\% & 4.39s     & 47\% & 1.04s\\
Shoulder    & 46\% & 5.61s & 3\%  & 2.16s     & 68\% & 1.77s\\
Elbow       & 13\% & 2.40s & 9\%  & 1.14s     & 45\% & 1.06s\\
Hand        & 24\% & 1.50s & 14\% & 0.98s     & 28\% & 0.86s\\
Hip         & 53\% & 7.28s & 99\% & 207.27s   & 65\% & 1.44s\\
Knee        & 59\% & 3.74s & 81\% & 12.2s     & 35\% & 0.78s\\
Foot        & 24\% & 0.78s & 23\% & 2.17s     & 41\% & 0.67s\\
All         & 27\% & 2.00s & 22\% & 1.70s     & 35\% & 0.95s\\
\hline\thickhline
\end{tabular}
}
\beforetab
\caption{Comparisons of our dataset with public human body-hand motion datasets. The occlusion and contact ratios are computed based on the algorithm introduced in Section~\ref{sec:occlusion_simulation}.
}
\aftertab
\label{tab:dataset_stats}
\end{table*}


We recruited 13 adult participants (10 male, 3 female) comfortable wearing a motion capture suit to go through sessions that typically took 1.5 hours each. At the beginning of each session, the participant changed into a velcro-covered mocap suit; a researcher attached 57 retro-reflective optical markers based on the Optitrack Biomech configuration; the participant then donned a pair of NANSENSE BioMed gloves.
We first captured calibration sequences for hands and body that consisted of exercising each joint independently. Next, we collected a few ``warm-up'' motions: idling, walking, crouching, jumping, talking on a phone, giving a thumbs up. Afterward, we focus on the collision- and contact-heavy ``social'' poses that are frequent in meetings. 
To capture a mix of sitting and standing poses, we used a stool and a table that could be rolled in and out of the session.  We brought the table in mid-way through the capture session and adjusted its height to be comfortable for the subject.  The table used a single leg to minimize occlusion and was heavily taped to reduce reflections. Toward the end of the session, the participant would take part in a two-to-five-minute unscripted conversation with the researcher, cued by a question; \eg, "What did you do over the weekend?" or "What was the last movie you watched?"




We found that videos were an effective way to convey to participants what types of motions to perform.  We studied CSPAN public domain videos from the United States Congress.  From these, we collected a set of 52 short segments to present to participants.  People in these clips perform a wide-ranging set of motions, including: resting their chins on their wrists, scratching their head or their ears, interlacing their fingers, gesturing, and crossing their arms.  The full set of videos can be seen in the supplemental material.
In an early internal test, we found that asking participants to mimic particular motions resulted in stiff, unnatural movement.  Therefore, we instead asked participants to use the clips only as inspiration.  Subjects used the videos in different ways, depending on their preference: some would watch the video as they performed; some would watch the video once and then come up with their own poses; and some preferred asking the researcher to demonstrate the poses.  In all cases, we emphasized that participants try to be as relaxed and natural as possible and take as much time as necessary to complete the capture.  

\section{Partial body pose prediction}
\label{sec:method}

\begin{figure*}[pt]
\centering
\includegraphics[width=0.8\textwidth]{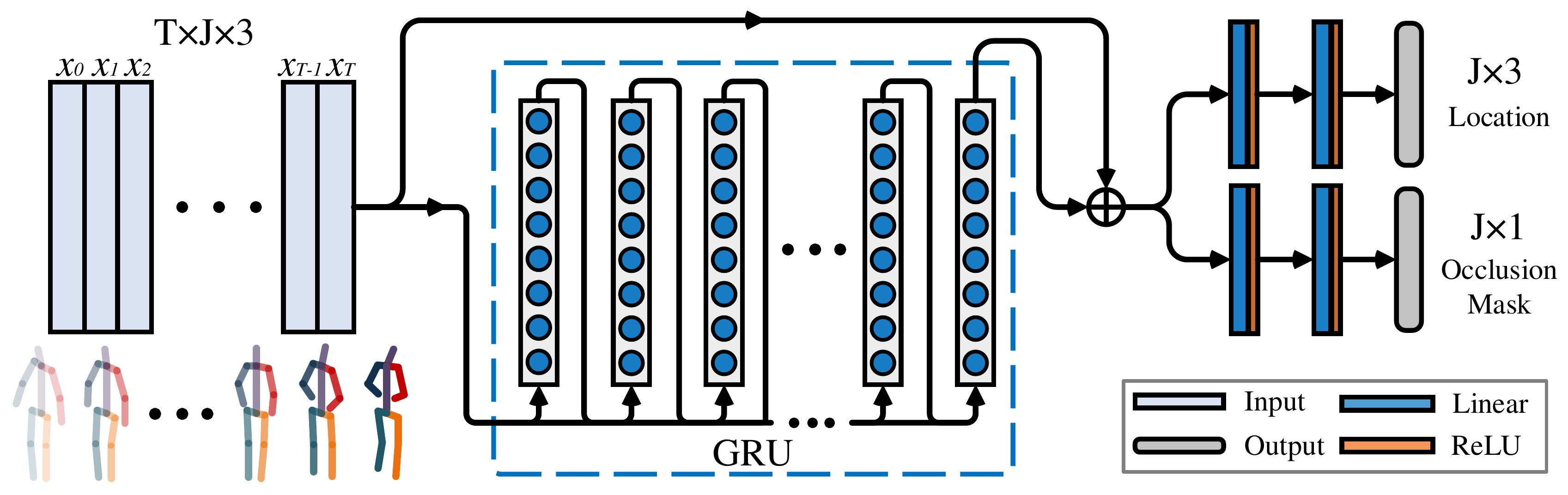}
\beforefigcaption
\caption{{\bf Proposed architecture}: We feed the input 3D joint locations into the gated recurrent units (blue box).  The output is then fed through fully connected layers to predict 3D joint positions (top branch) and whether the joint from the input is occluded (bottom branch).}
\afterfigcaption
\label{fig:model}
\end{figure*}

In this section, we discuss our method for predicting the full body pose when some joints are occluded or out of the camera frustum. 

\beforesubsection
\subsection{Occlusion simulation}
\label{sec:occlusion_simulation}
\aftersubsection

To evaluate our network on its capability to reconstruct untracked joints, we annotate the joint locations in the ground truth data with a label specifying if a given joint is occluded or outside the camera frustum.  To simulate the multiple inside-out tracking cameras of headsets like the Oculus Quest or Microsoft Hololens, we place a virtual camera four centimeters in front of the nose.
The camera uses a circular field-of-view of 200$^\circ$, and is tilted downwards by 15$^\circ$.  Joints are tagged "out-of-view" if they are outside this cone.  In addition, joints closer than 12cm to the camera center are considered untracked because they would typically be out of focus and lack visual context.

\begin{figure}[ptb]
\centering
\includegraphics[width=\linewidth]{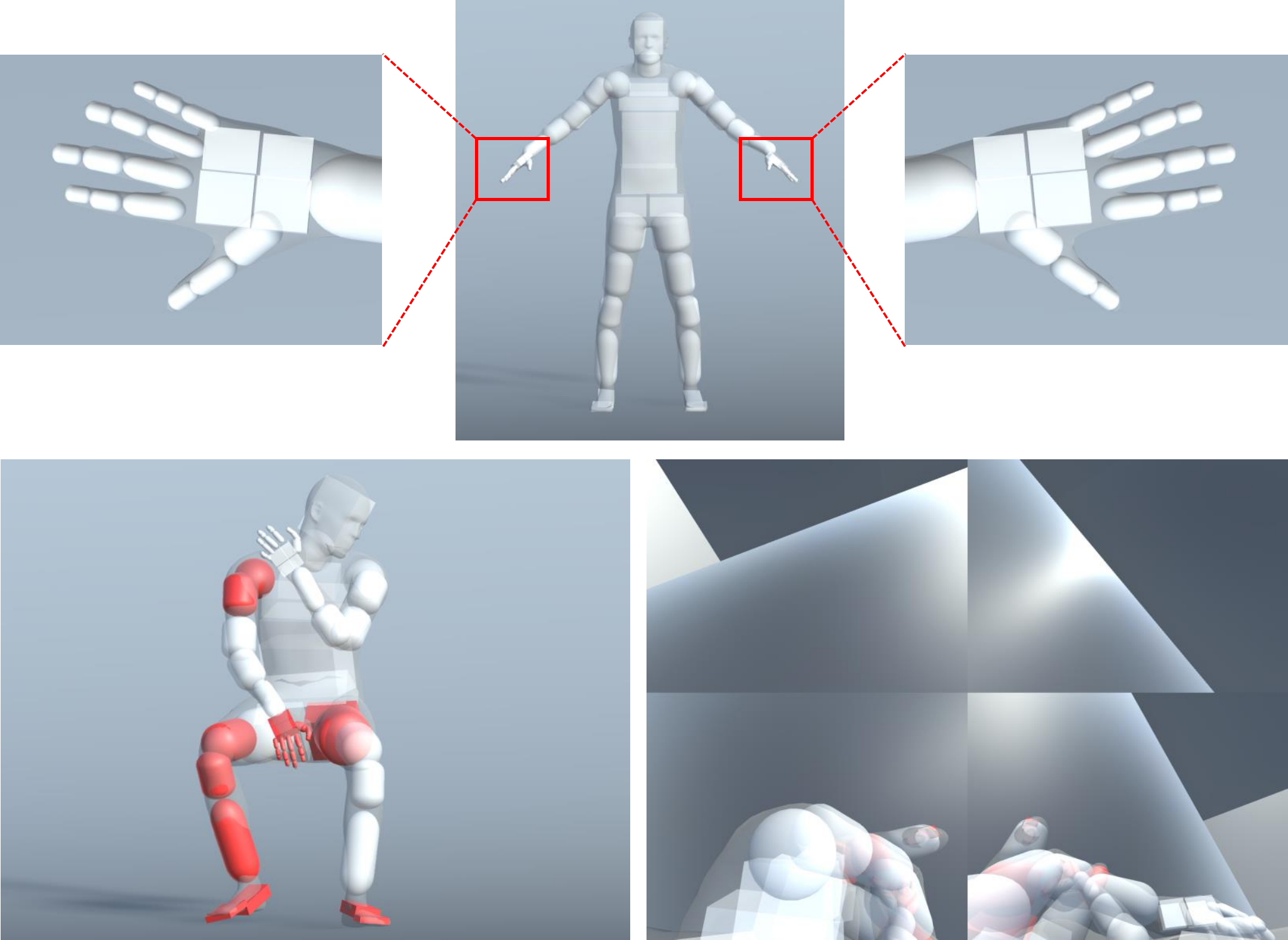}
\beforefigcaption
\caption{Our occlusion simulation uses geometric primitives (first row) as proxies for body part visibility. We show an example pose (second row, left) with detected occlusions in red. Occlusions are detected using a simulated camera with 200$^\circ$ field of view (approximated field of view using 4 cameras in (second row, right)) virtually attached to the front of a VR headset.}
\afterfigcaption
\label{fig:occ_sim}
\end{figure}

To simulate occlusion efficiently, we use a simplified geometric model where each bone segment is represented by a capsule or bounding box~\cite{CollisionDetection}.  We test visibility by tracing rays from the camera center and check whether they intersect any primitive before reaching the skeleton.  A given body joint is considered visible if the joint itself is visible as well as at least one primitive attached to the joint (see Fig.~\ref{fig:occ_sim}). 
In our experiments, this heuristic works well and requires less fine tuning than using an image-based occlusion simulation (i.e. counting visible pixels per body part, or training a neural network on our inside-out rendered views).

In this manner, we test for the visibility of the shoulders, elbows, hips, and knees.  We treat the feet specially, as the ankle joint itself is almost always occluded; instead, all the joints in the feet are considered visible if any (including the toes) is visible from the camera.  A similar procedure applies to the hand: while fingers frequently self-occlude, evidence from the vision-based tracking literature suggests it is possible to predict a reasonable hand pose even when some digits are hidden~\cite{MultiModalHandMocap20}.  We therefore mark the entire hand as visible provided at least 65\% of its primitives are visible.  

\beforesubsection
\subsection{Embodied pose prediction}
\label{sec:prediction}
\aftersubsection

Given these occlusion annotations, we can define the partial pose prediction problem.  Given the 3D locations of all the joints that are {\em tracked}, we will predict the 3D locations of {\em all} the body joints.  While we focus here on 3D joint locations, these can be post-processed into a fully posed human body model using an inverse kinematics (IK) solver~\cite{Stoll2011,Parger2018}.

It is often intractable to predict missing joints from single-frame inputs. Motivated by~\cite{pavllo2019,Cai2019}, we incorporate the history and tackle the problem through a temporal model.
While extrapolation based on constant position velocity is a reasonable baseline~\cite{Martinez2017}, the best extrapolation is likely situation-dependent: a palm supporting a cheek should be pinned relative to the head pose, while a gesturing hand might finish resting on a hip.  Therefore, we train models that {\em learn} to extrapolate from partially observed present and historical joint locations.    

\paragraph{Proposed architecture} Our network takes the last 27 tracked frames (at 30fps, this results in a receptive field of 0.9 seconds) as input (\ie, time window size T=27). For occluded or out-of-view joints, we repeat the last {\em known} position.
Our network backbone uses Gated Recurrent Units (GRU)~\cite{GRU} with a 512-dimensional hidden state size, which formulates the sequential data into a hidden Markov Chain model and describes relations among neighboring nodes with sharing weights.

For a node $t$ representing the $t$-th frame in the sequence, we use $t-1$ and $t+1$ to denote the previous frame and the next frame, respectively. Given an input feature encoding $x_t$ for node $t$, the output $y_t$ is determined jointly by the hidden states (\ie, linear weights $W$ and bias term $b$), update gate $z$ and reset gate $r$:
\begin{equation}\begin{aligned}
z_t &= \sigma(\,W_{z} x_{t} + U_{z} y_{t-1} + b_z\,), \\
r_t &= \sigma(\,W_{r} x_{t} + U_{r} y_{t-1} + b_r\,), \\
\hat{y}_t &= \phi(\,W_{y} x_{t} + U_{y}( r_{t} \odot y_{t-1} ) + b_y\,),\\
y_t &= (1 - z_t) \odot y_{t-1} + z_t \odot \hat{y}_t,
\end{aligned}\end{equation}
where $\hat{y}$ denotes the candidate activation, $\odot$ denotes the Hadamard product, $\sigma$ denotes the sigmoid function and $\phi$ denotes the hyperbolic tangent function. The model takes the (partially occluded) 3D joint locations as input, and the result of the GRU is fed through two branches of fully connected layers to predict 3D positions and occlusion labels, respectively (see Fig.~\ref{fig:model}). 

In an egocentric setting, the 3D body joint positions are given as offsets from the head location, while the fingers are given as wrist-local joint positions, removing inherited inaccuracies from the wrist position.
Our experiments show that this helps the network to predict small nuances in gestures and finger poses, without being penalized for the global positional error (see Table~\ref{tab:body_part_pred_ablation}).
We ask our network to {\em predict} whether a given joint is occluded, rather than giving the occlusion mask as input to encourage it to model occlusion.
This has similar accuracy to providing the occlusion mask as input, but the transitions between visible and occluded joints in the prediction are noticeably smoother.  
The percentage of frames exhibiting strong acceleration ($> 9 m/s^2$) is reduced from 6.1\% to 3.5\% by predicting the occlusion mask (compared to 0.2\% in the ground truth).
In the final architecture, the output of the GRU is concatenated with the input through skip connections and branched out to two MLPs with the number of hidden neurons $\{512, 512\}$ for two tasks: joint location prediction and occlusion classification. 
\newcommand{\Lpos}[0]{\ensuremath{L_{\text{pos}}}}
\newcommand{\Lkin}[0]{\ensuremath{L_{\text{kin}}}}
\newcommand{\Locc}[0]{\ensuremath{L_{\text{occ}}}}
\newcommand{\euclideanDistance}[2]{D(#1,\, #2)}
\newcommand{\crossentropy}[2]{H(#1,\, #2)}

\paragraph{Losses and training} To train the network, we use a weighted sum of multiple losses. The total loss function is defined as
\begin{equation}
L = \Lpos{} + \Lkin{} + \Locc{},
\end{equation}
where \Lpos{} denotes the global position loss, \Lkin{} the kinematic loss, and \Locc{} the occlusion mask loss. For the global position loss and the kinematic loss, we measure them by the Euclidean distance $\euclideanDistance{\cdot}{\cdot}$.

\Lpos{} measures the distances between the ground truth joint position $\hat{p}_i$ and the predicted joint position $p_{i}$ in the head/wrist-local frame as described above. We weigh finger joint errors higher than body joint, and occluded joints higher than visible, \ie,
\begin{equation}
\Lpos = \frac{1}{J}\sum\limits_i^J \euclideanDistance{p_i}{\hat{p}_i}.
\end{equation}

\Lkin{} penalizes violations of kinematic limits of the skeleton, helping to ensure that the bones $\{b_{i,j} = p_i - p_j\}$ are valid in the predicted pose. The first term measures the differences on the parent-local joint positions $b_i$ and the second term measures errors in bone lengths, that is,
\begin{equation}
\Lkin = \sum\limits_{i}^J \sum\limits_{j\in ch(i)} \euclideanDistance{b_{i,j}}{\hat{b}_{i,j}} + \euclideanDistance{\|b_{i,j}\|_2}{\|\hat{b}_{i,j}\|_2},
\end{equation}
where $ch(i)$ denotes the children joints of joint $i$ in the kinematics tree.

\Locc{} penalizes the misclassification of the predicted occlusion labels against ground-truth, \ie,
\begin{equation}
\Locc = \sum\limits_{i}^J \crossentropy{o_i}{\hat{o}_i},
\end{equation}
where $\crossentropy{}{}$ denotes the cross entropy for the binary classification.

In our experiments, we use a weight of $10$ for the head-local body joint position loss \Lpos{}, and $100$ for the wrist-local finger joint position loss, both weights are increased by $20\%$ if the joint is occluded. 
We weigh the parent-local position loss by $3$, which is also increased by $20\%$ for occluded joints.  
We weigh the bone length errors of body joints by $7$ and fingers by $10$ for \Lkin{}. 
The weight for the occlusion mask loss \Locc{} is set to $0.025$.  We implement our method in PyTorch and train it using the Adam optimizer with a learning rate starting from 1e-3. The models are trained over 5 epochs, with a batch size schedule of $\{256, 512, 1024, 1024, 1024\}$. 

\paragraph{Post-processing}  3D full-body avatars are typically animated using {\em forward kinematics}, where the pose is encoded using local joint angles for each bone relative to its parent.  Because the output of our networks are 3D world-space joint positions, computing the pose requires solving for the body pose using {\em inverse kinematics (IK)}.
Our IK solver computes the body joint angles that respect the hard bone-length constraints while minimizing the world-space positions' squared error by the Gauss-Newton algorithm, initialized from a rest pose for the first frame, and the previous frame subsequently. We use fixed-offset auxiliary joints at the end of the kinematic chain (e.g. finger tips) to disambiguate joint orientations.
When joints become visible after being occluded for a while, the neural network snaps the joints to their new location which results in fast, unrealistic motions.
We experimented with temporally consistent networks, however these networks traded accuracy for smoothness.
Using a simple momentum-based algebraic smoothing instead, we achieve very smooth motion without sacrificing accuracy (see Section~\ref{sec/eval/io-bt}).

\begin{figure*}
\centering
\resizebox{0.9\linewidth}{!}{
\begin{tabular}{@{}ccccc||ccccc@{}}
    \subfloat{\includegraphics[width=0.10\linewidth]{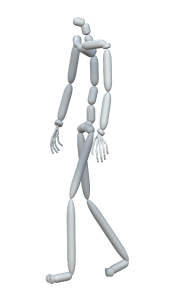}} &
    \subfloat{\includegraphics[width=0.10\linewidth]{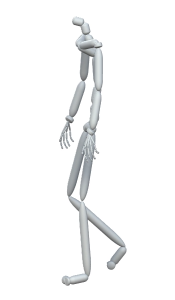}} &
    \subfloat{\includegraphics[width=0.10\linewidth]{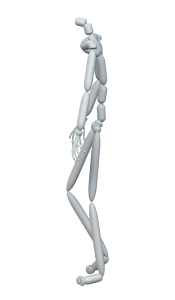}} &
    \subfloat{\includegraphics[width=0.10\linewidth]{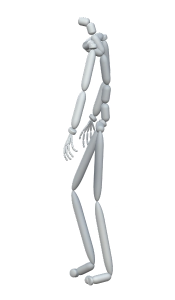}} & 
    \subfloat{\includegraphics[width=0.10\linewidth]{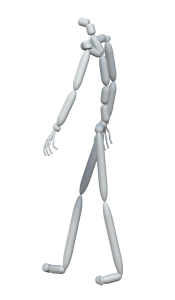}} \hspace{3mm} & \hspace{3mm}
    \subfloat{\includegraphics[width=0.10\linewidth]{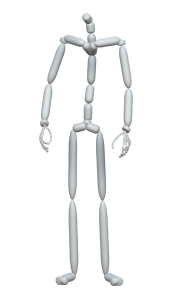}} &
    \subfloat{\includegraphics[width=0.10\linewidth]{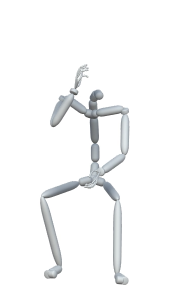}} &
    \subfloat{\includegraphics[width=0.10\linewidth]{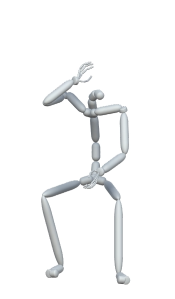}} &
    \subfloat{\includegraphics[width=0.10\linewidth]{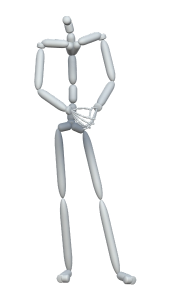}} &
    \subfloat{\includegraphics[width=0.10\linewidth]{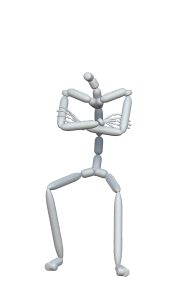}}
\\
    \subfloat{\includegraphics[width=0.10\linewidth]{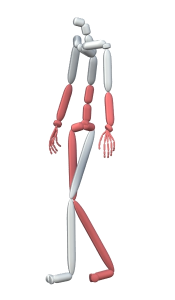}} &
    \subfloat{\includegraphics[width=0.10\linewidth]{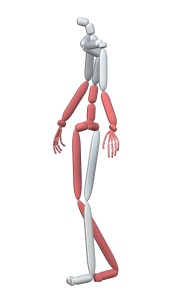}} &
    \subfloat{\includegraphics[width=0.10\linewidth]{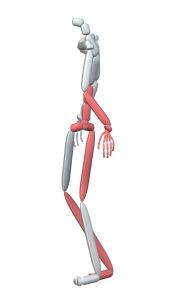}} &
    \subfloat{\includegraphics[width=0.10\linewidth]{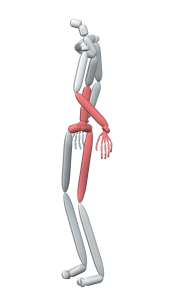}} &
    \subfloat{\includegraphics[width=0.10\linewidth]{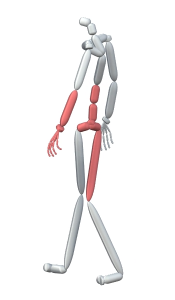}} \hspace{3mm} & \hspace{3mm}
    \subfloat{\includegraphics[width=0.10\linewidth]{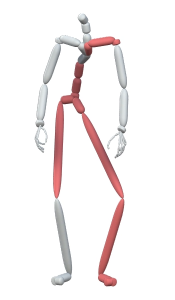}} &
    \subfloat{\includegraphics[width=0.10\linewidth]{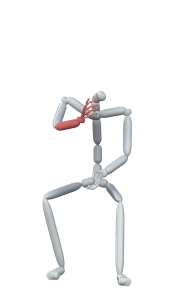}} &
    \subfloat{\includegraphics[width=0.10\linewidth]{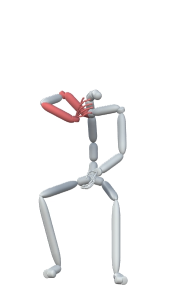}} &
    \subfloat{\includegraphics[width=0.10\linewidth]{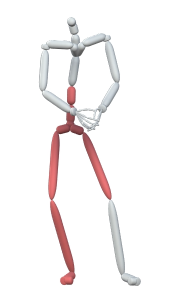}} &
    \subfloat{\includegraphics[width=0.10\linewidth]{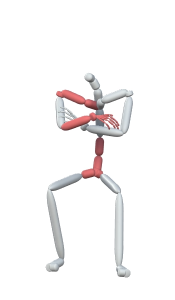}}
\\
    \subfloat{\includegraphics[width=0.10\linewidth]{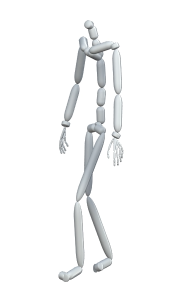}} &
    \subfloat{\includegraphics[width=0.10\linewidth]{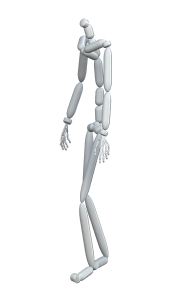}} &
    \subfloat{\includegraphics[width=0.10\linewidth]{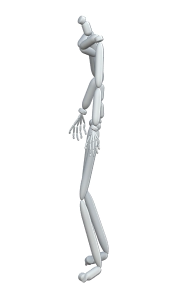}} &
    \subfloat{\includegraphics[width=0.10\linewidth]{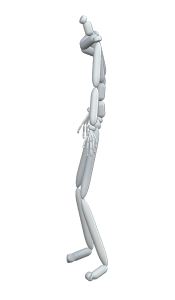}} &
    \subfloat{\includegraphics[width=0.10\linewidth]{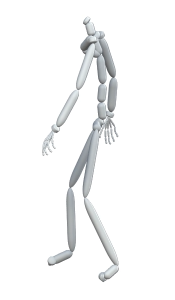}} \hspace{3mm} & \hspace{3mm}
    \subfloat{\includegraphics[width=0.10\linewidth]{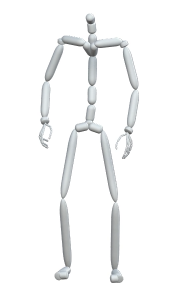}} &
    \subfloat{\includegraphics[width=0.10\linewidth]{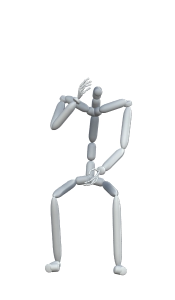}} &
    \subfloat{\includegraphics[width=0.10\linewidth]{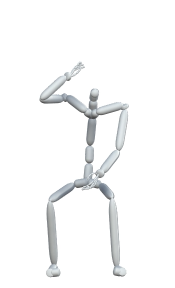}} &
    \subfloat{\includegraphics[width=0.10\linewidth]{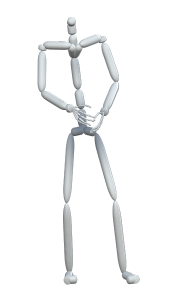}} &
    \subfloat{\includegraphics[width=0.10\linewidth]{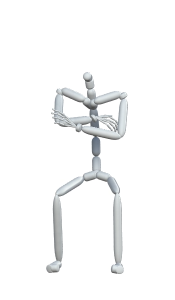}}
\\
    \subfloat{\includegraphics[width=0.10\linewidth]{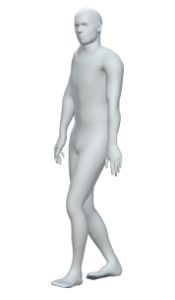}} &
    \subfloat{\includegraphics[width=0.10\linewidth]{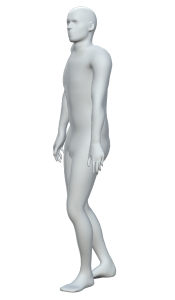}} &
    \subfloat{\includegraphics[width=0.10\linewidth]{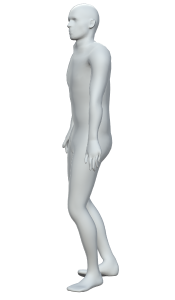}} &
    \subfloat{\includegraphics[width=0.10\linewidth]{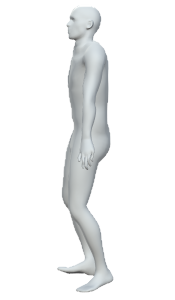}} &
    \subfloat{\includegraphics[width=0.10\linewidth]{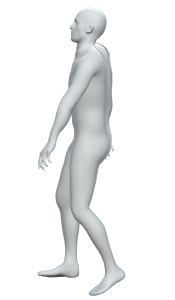}} \hspace{3mm} & \hspace{3mm}
    \subfloat{\includegraphics[width=0.10\linewidth]{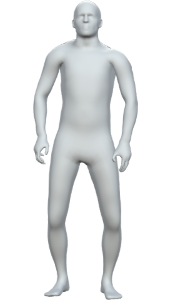}} &
    \subfloat{\includegraphics[width=0.10\linewidth]{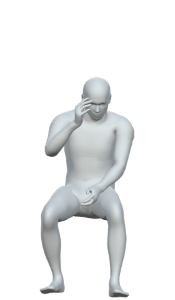}} &
    \subfloat{\includegraphics[width=0.10\linewidth]{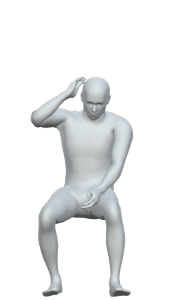}} &
    \subfloat{\includegraphics[width=0.10\linewidth]{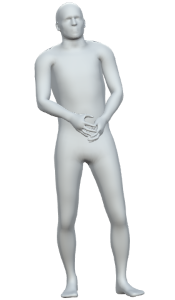}}  &
    \subfloat{\includegraphics[width=0.10\linewidth]{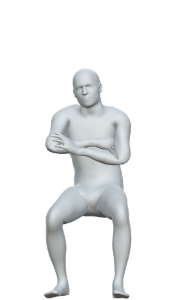}}
\end{tabular}
}
\beforefigcaption
\caption{Qualitative results of our method on \dataset. From top to bottom, we show ground truth poses, last tracked joint positions, predicted poses with our model and IK solved avatars from our predicted poses.  Red areas show the significant occlusion from the egocentric view, which we are able to predict with high accuracy.
}
\afterfigcaption
\label{fig:body_pred}
\end{figure*}

\section{Evaluation}

We evaluate our proposed method and the \dataset dataset on three applications.  First, we compare inside-out body tracking using both the occlusion simulation and occlusion prediction method described in Section~\ref{sec:method}.  We show that our proposed method generates smooth and plausible motion for this application, and that our dataset is critical for covering new motions for embodied presence.
Because inside-out body tracking is a nascent area where camera placement and datasets have not been standardized, we also show that our method is effective on two additional applications: 3-point upper body tracking and finger synthesis.  In these tasks, the missing information does not vary across frames, so we eliminate the occlusion mask prediction branch of our method. For all applications, we use an 80\%/20\% train/test split and ensure that the test set includes two participants not seen during training. We use root mean square per-joint position error (RMSJPE) and mean per-joint position error (MPJPE) as the evaluation metric.  All units are in centimeters.

\begin{table*}[ptb]
\centering
\setlength{\tabcolsep}{6pt}
\renewcommand\arraystretch{1.2}
\resizebox{0.9\linewidth}{!}{
\begin{tabular}{c|cc|cc|cc|cc|cc}
\hline\thickhline
\multirow{2}{*}{Body Part} & 
\multicolumn{2}{c|}{GRU (Ours)} & 
\multicolumn{2}{c|}{Baseline} & 
\multicolumn{2}{c|}{RNN} & 
\multicolumn{2}{c|}{LSTM} & 
\multicolumn{2}{c}{TCN} \\
\cline{2-11}
& Occluded & All & Occluded & All & Occluded & All & Occluded & All & Occluded & All \\
\cline{1-11}
Body    & \bf{5.9/4.0} & \bf{4.3/3.1}   & 13.1/8.5 & 7.4/2.7    & 6.1/4.2 & 4.6/3.3     & 6.4/4.2 & 4.5/3.2     & 6.2/4.2 & 4.9/3.7 \\
Finger  & \bf{4.1/3.0} & \bf{2.0/1.2}   & 5.0/3.2 & 2.2/0.6     & 4.4/3.2 & 2.2/1.3     & 4.3/3.0 & 2.1/1.3     & 4.4/3.2 & 2.2/1.4 \\
Body \& Finger  & \bf{9.3/5.9} & \bf{5.3/3.3}    & 15.9/9.6 & 7.6/2.2    & 10.1/6.3 & 5.9/3.7   & 10.2/6.1 & 5.8/3.6    & 10.3/6.4 & 6.2/4.1 \\
\hline\thickhline
\end{tabular}
}
\beforetab
\caption{Comparisons for inside-out body tracking on \dataset dataset. We use RMSJPE/MPJPE to measure tracking errors of \textit{occluded/all} joints in cm. \textit{Body \& Finger} and \textit{body} errors are measured in the global frame while finger errors are measured in the wrist-local frame.
}
\aftertab
\label{tab:body_part_pred_ablation}
\end{table*}

\begin{table*}[ptb]
\centering
\setlength{\tabcolsep}{6pt}
\renewcommand\arraystretch{1.2}
\resizebox{\linewidth}{!}{
\begin{tabular}{c|cc|cc|cc|cc|cc|cc}
\hline\thickhline
\multirow{2}{*}{Body Part} & 
\multicolumn{2}{c|}{GRU (Ours)} & 
\multicolumn{2}{c|}{GRU \textdagger} & 
\multicolumn{2}{c|}{GRU *} & 
\multicolumn{2}{c|}{GRU 9F} &
\multicolumn{2}{c|}{GRU 81F} & 
\multicolumn{2}{c}{GRU 1024} \\
\cline{2-13}
& Occluded & All & Occluded & All & Occluded & All & Occluded & All & Occluded & All & Occluded & All \\
\hline
Body    & \bf{5.9/4.0} & \bf{4.3/3.1}   & 6.4/4.6 & 4.7/3.5     & 6.0/4.2 & 4.5/3.4   & 6.1/4.2 & 4.4/3.2   & 6.3/4.3 & 4.6/3.3     & 5.8/3.9 & 4.2/3.0   \\
Finger  & \bf{4.1/3.0} & \bf{2.0/1.2}   & 4.4/3.5 & 2.7/2.2     & 4.1/3.0 & 2.1/1.3   & 4.2/3.0 & 2.0/1.2   & 4.3/3.0 & 2.1/1.2     & 4.3/3.0 & 2.0/1.2   \\
Body \& Finger & \bf{9.3/5.9} & \bf{5.3/3.3}    & 9.3/6.1 & 5.3/3.3    & 9.5/6.2 & 5.7/3.7    & 9.5/6.0 & 5.5/3.5   & 10.3/6.2 & 5.8/3.6     & 9.3/5.9 & 5.3/3.3     \\
\hline\thickhline
\end{tabular}
}
\beforetab
\caption{Comparisons for inside-out body tracking on \dataset dataset. \textdagger~ uses head-local fingers. * is without skip connections. 9F and 81F refer to using 9 and 81 frames of as network input instead of 27.
}
\aftertab
\label{tab:body_part_pred_ablation_params}
\end{table*}

\begin{table*}[ptb]
\centering
\setlength{\tabcolsep}{8pt}
\renewcommand\arraystretch{1.2}
\resizebox{\linewidth}{!}{
\begin{tabular}{c|c|c|c|c|c|c|c|c}
\hline\thickhline
& \multicolumn{4}{c|}{Global position error} & \multicolumn{4}{c}{Parent local position error} \\
\cline{2-9}
& Occluded & $\Delta$ & All & $\Delta$ & Body-only & $\Delta$ & All & $\Delta$ \\
\hline
Ours                        & \bf{5.9/4.0} &      & \bf{4.3/3.1}  &         & \bf{0.90/0.61} &        & \bf{0.51/0.34} & \\
w/o occlusion prediction    & 6.2/4.2 & +5\%/+5\% & 4.4/3.2 & +2\%/+2\%     & 0.92/0.63 & +2\%/+3\%   & 0.54/0.37 & +6\%/+9\% \\
w/o parent local            & 6.0/4.1 & +2\%/+3\% & 4.4/3.2 & +2\%/+2\%     & 0.91/0.65 & +1\%/+7\%   & 0.53/0.37 & +4\%/+9\% \\
w/o bone length             & 6.0/4.0 & +2\%/+0\% & 4.3/3.1 & +0\%/+0\%     & 1.06/0.71 & +18\%/+16\% & 0.59/0.39 & +16\%/+15\% \\
w/o par. loc. \& bone len.  & 6.0/4.0 & +2\%/+0\% & 4.4/3.2 & +2\%/+2\%     & 1.17/0.79 & +30\%/+30\% & 0.62/0.39 & +22\%/+15\% \\
\hline\thickhline
\end{tabular}
}
\beforetab
\caption{Ablation study of losses for inside-out body tracking on \dataset dataset. We use RMSJPE/MPJPE to measure tracking errors of \textit{occluded/body/all} joints in cm. Parent local error can be understood as bone length error.
}
\aftertab
\label{tab:body_part_pred_loss_ablation}
\end{table*}

\subsection{Inside-out body tracking}
\label{sec/eval/io-bt}

We first evaluate our proposed body prediction method (see Section~\ref{sec:prediction}) on three datasets:  our own \dataset dataset, Mixamo~\cite{Xu2019DenseRaC} and TWH16.2M~\cite{Lee2019}.  
For each dataset, we first simulate the output of an inside-out body tracking system using the occlusion simulation step described in Section~\ref{sec:occlusion_simulation} and then train a model to evaluate the tracking accuracy. 
To the best of our knowledge, there is no other published method that is applicable to this complex missing data problem.
We experimented with applying IK solutions and smoothing for the baseline (see the user study description), which increased the overall error.
To this end, we set up a baseline, which copies the last tracked position for occluded joints while using the ground truth for tracked joints.
As shown in Table~\ref{tab:body_part_pred_ablation} (first two columns) and Fig.~\ref{fig:body_pred}, our method is able to significantly reduce the joint positional error over the baseline model.
For occluded body joints, our methods approximately halves the error; for occluded fingers we reduce it by nearly one centimeter. In Table~\ref{tab:cross_eval} (left), we show how our model trained on one dataset performs when predicting body poses on other datasets. It can be observed that our dataset is unique and the set of motions that we use in our animations are not covered by other datasets.

\paragraph{Ablation studies} We conduct an ablation study of different model architectures and representations, as shown in Table~\ref{tab:body_part_pred_ablation} and~\ref{tab:body_part_pred_ablation_params}. We first try replacing GRU with a few other popular sequential neural networks (RNN, LSTM~\cite{LSTM}, TCN~\cite{pavllo2019}). For a fair comparison, we use the same hidden state size as our proposed architecture for RNN and LSTM and we use the causal convolution version for TCN to be consistent with our online inference setting. From the results, it can be observed that GRU performs best in our experiment setting compared to other temporal neural networks. We further evaluate several different variants of the proposed network architecture, \ie, with/without skip connections, different coordinate system representations, different time window sizes and different hidden state sizes.
From the results, we find that 1) adding skip connections helps preserve visible joint positions and generates smoother results, 2) wrist-local finger coordinates improve accuracy over head-local (\ie, our general body representation) and 3) the current hyper-parameter configuration obtains fairly good performance.

Furthermore, we conduct an ablation study of different loss combinations, as well as when removing the occlusion prediction branch, see Table~\ref{tab:body_part_pred_loss_ablation}. The results show that predicting the occlusion masks improves the joint accuracy of occluded joints. Removing the parent local position error loss, the bone length loss, or both at once, greatly increases the bone length errors compared to using both losses together.

\newcommand{\baseline}{\textsc{Baseline}}
\newcommand{\ourscond}{\textsc{Ours}}
\newcommand{\gtcond}{\textsc{GT}}

\begin{figure*}[ptb]
\centering
\includegraphics[width=0.333\linewidth]{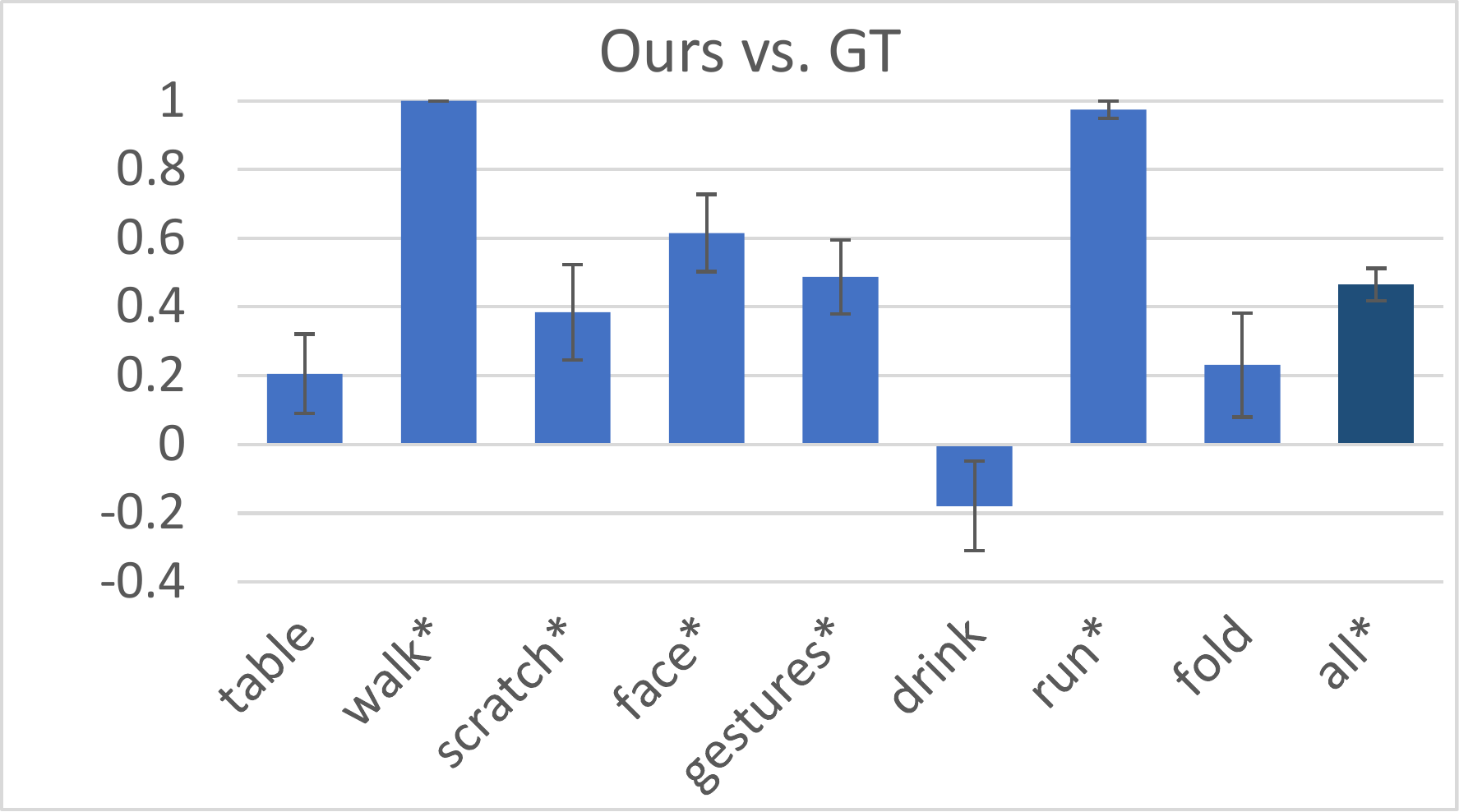}\hspace{-1mm}
\includegraphics[width=0.333\linewidth]{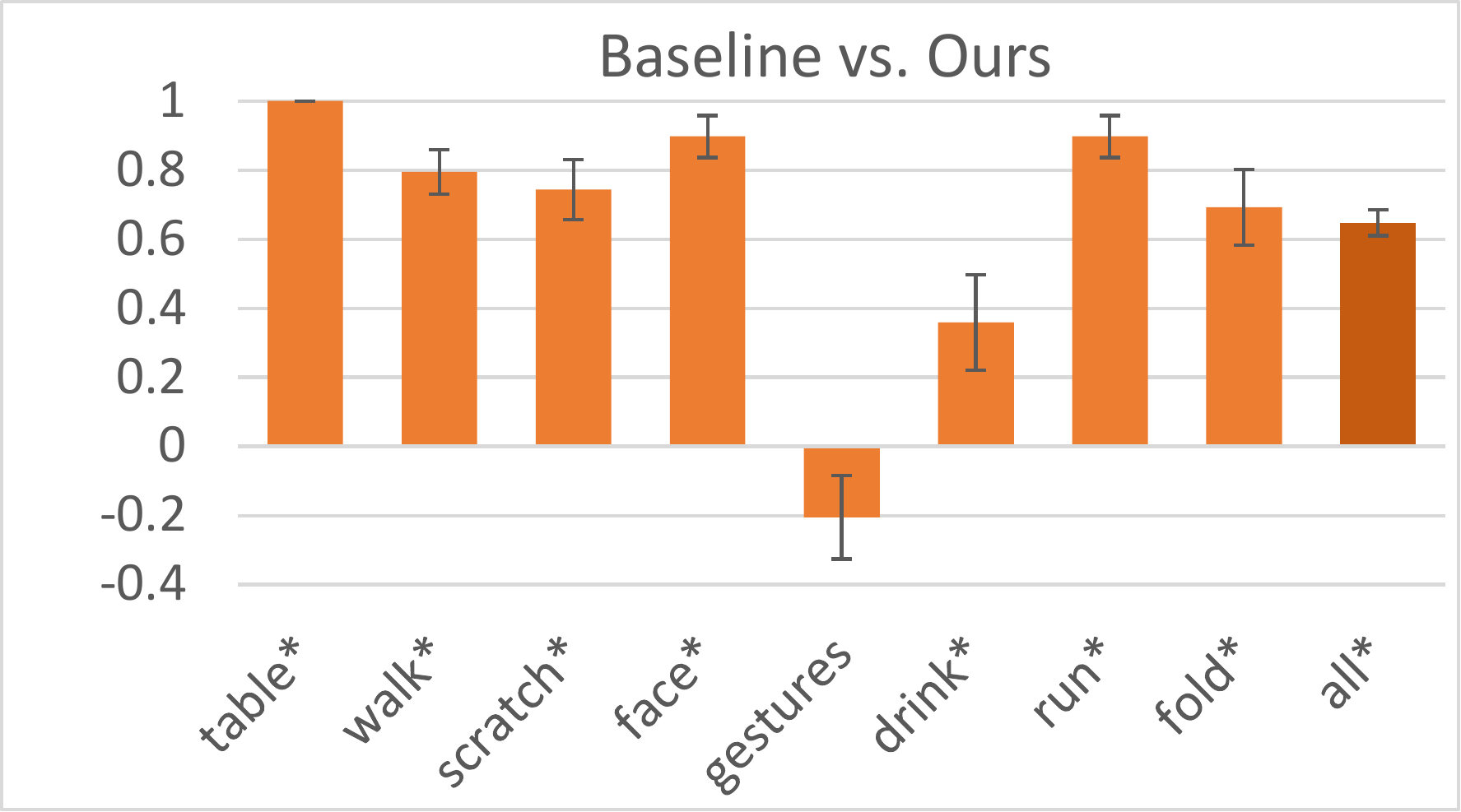}\hspace{-1mm}
\includegraphics[width=0.333\linewidth]{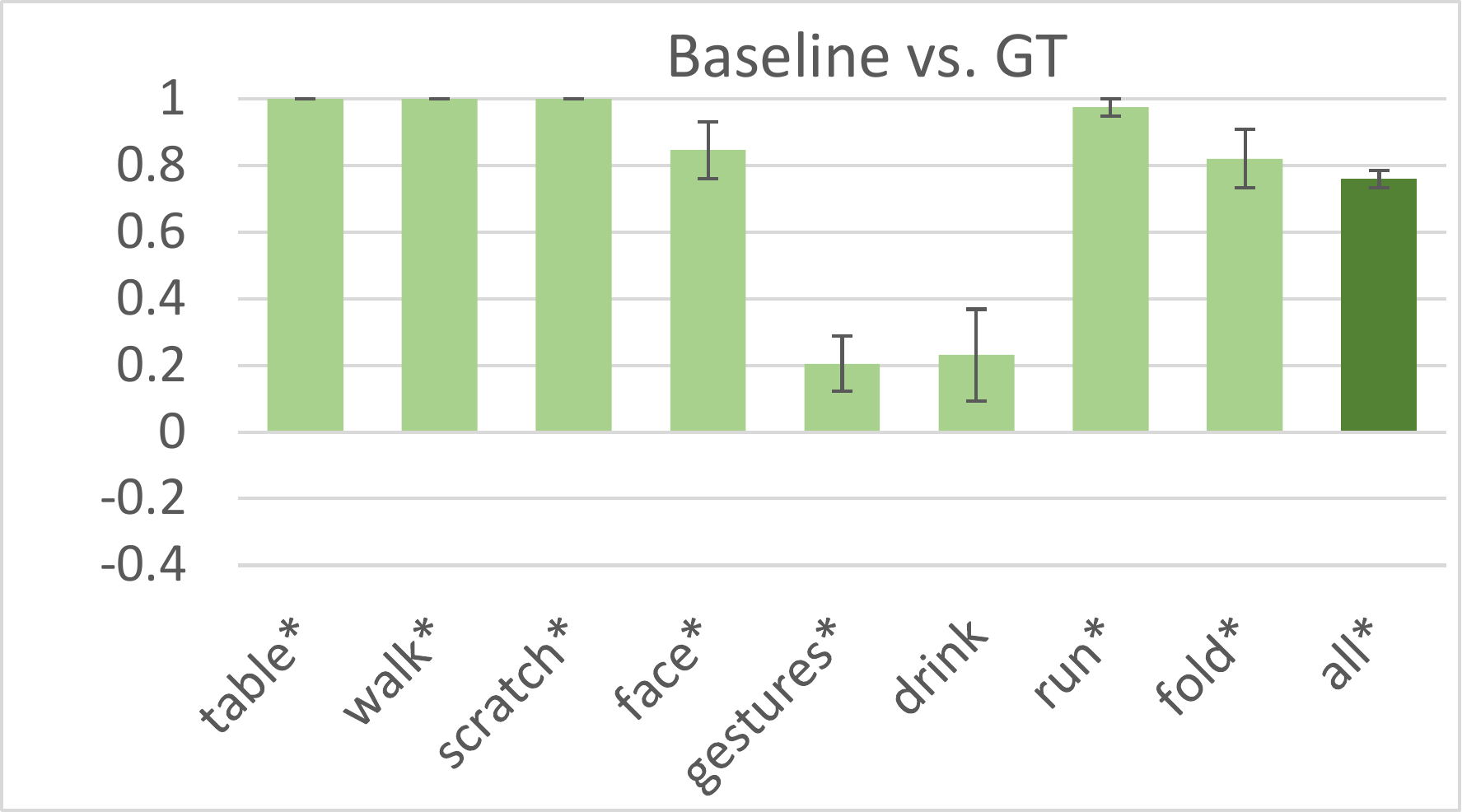}
\caption{Results of our user study. Values $>0$ show a preference of the second technique over the first, i.e. \gtcond~over \ourscond, \ourscond~over \baseline~and \gtcond~over \baseline. Error bars show standard error.  $^*$ indicates statistical significant difference between the methods according to a t-test (all significant results achieved at least $p<.01$ except for \emph{gestures} for \baseline~vs. \gtcond~with  $p<.02$).}
\label{fig:study-results}
\end{figure*}

\paragraph{User study}


To assess the practical value of our approach, we conducted a user study.
Due to COVID-19, we opted for a video comparison study rather than an actual VR study.
From our test data set, we randomly selected eight sequences with varying motions and levels of occlusion, including: leaning on a table, walking, scratching the head, touching the face, finger gestures, drinking from a glass, running, and crossing arms.
For each of these animations, we created renderings with our standard avatar using three different methods: the fully captured ground-truth (\gtcond{}), using the last-known position for each joint (\baseline{}), and using the predictions from our trained network (\ourscond{}).  In both of the test conditions (\baseline{} and \ourscond{}), rendering the body mesh requires converting from 3D positions to joint angles, so we use the IK solver described in Section~\ref{sec:prediction} (configured identically for both cases) and apply the same momentum-based post-solve smoothing. 
After the smoothing and the IK solver stage, the root mean square joint position error in \ourscond~decreases by $0.04cm$ for occluded body joints, while the overall error increases by $1.5cm$ compared to the network output positions.
The root mean square joint position error of {\em occluded} body joints in the \baseline{} was reduced by $3.5cm$, while the overall error was increased by $2.4cm$ using the smoothed, IK solved positions.

In our study, we showed side-by-side views of all pairs of renderings to the participants, following a ``similarity judgment'' design\cite{mantiuk2012comparison}.
For each pair, the participants were asked to choose the more natural result (``Which animation is more natural?''). They could choose either video or indicate that they were equally natural.
To avoid biases and learning effects, we randomized the order of pairs as well as the position of the videos while ensuring that for each participant the same number of techniques were shown on the left and right.
To evaluate the results, we attribute the choice of one technique with $+1$ and the other with $-1$.
Averaging over all eight scenes results in a quality score for each pair of techniques.
Using a t-test, we determine the probability of the drawn sample to come from a zero-mean distribution---zero-mean would indicate both techniques to be of equal quality.

We recruited 39 participants with medium to high experience with VR from a local university. Each pair-wise comparison video was combined into a single clip to allow for synchronous playback. The participants were told that they could replay and navigate in the video as they wished.

The results of the study are shown in Fig.~\ref{fig:study-results}. All three pair-wise comparisons showed significant results:
\ourscond{} vs. \gtcond{} (mean $=0.46$, std $=0.30$, $t(38)=-9.74$, $p<.001$), \baseline{} vs. \ourscond{} (mean $=0.65$, std $=0.23$, $t(38)=-17.50$, $p<.001$), and \baseline{} vs. \gtcond{} (mean $=0.76$, std $=0.16$, $t(38)=-29.07$, $p<.001$), with \gtcond{} beating \baseline{} and \ourscond{}, and \ourscond{} being significantly better than \baseline{}.


A detailed analysis of the individual animations further showed that our method achieves significantly better results than the baseline in seven out of eight cases. Only one animation with limited occlusions, where the avatar is standing straight and showing some hand signs (peace, thumbs up, etc.) could not achieve a significant improvement.  This might be due to a reduced accuracy in the hand pose: the predicted hand looks slightly closed compared to the tracked pose. Furthermore, the ground truth  achieved significantly better results than our model in only five out of eight cases. Two such cases are walking and running animations. In the six remaining sequences, containing body-body part contact and hand-object interaction, we achieve mean scores between -0.18 up to 0.61, three of which see a significant preference for the ground truth.

The results of the study show that a learned prior using our approach clearly creates more realistic animations compared to the baseline which uses only an IK solver and a smoothness term on the last tracked positions.
Post-study interviews revealed that the baseline animations were perceived as partially unnatural and jerky.
Some poses chosen by the baseline clearly resulted in ``uncomfortable poses humans would not choose freely.''
Furthermore, the animations generated by our solution were described as ``completely plausible'' and ``very natural''.
However, a direct comparison to ground truth animations still shows a quality gap.
This is not surprising though, as even with good priors, predicting hidden motions will not produce sufficient detail unless random animations are hallucinated even when the tracked body pose is not changing (e.g. fingers typing on a keyboard).
Participants tellingly described the differences as ``lacking detail'' or ``lacking depth.''
Still, overall our approach can clearly help provide realistic and natural animations for hidden joints.

\begin{table*}[ptb]
\centering
\setlength{\tabcolsep}{6pt}
\renewcommand\arraystretch{1.2}
\resizebox{0.5\linewidth}{!}{
\begin{tabular}{c|cccc}
\hline\thickhline
\diagbox{Test}{Train} 
& \dataset      & TWH16.2M      & Mixamo        
& Input \\
\hline
\dataset            & \bf{9.3 / 5.9}    & 15.6 / 10.3    & 13.2 / 8.2    
& 15.9 / 9.6    \\
TWH16.2M            & 8.2 / 6.1     & \bf{5.8 / 2.8}     & 10.3 / 8.3     
& 9.1 / 6.0     \\
Mixamo              & 22.4 / 14.6   & 28.8 / 19.2   & \bf{17.8 / 11.3}   
& 23.1 / 14.7   \\
\hline\thickhline
\end{tabular}}
\hspace{5mm}
\resizebox{0.38\linewidth}{!}{
\begin{tabular}{c|ccc}
\hline\thickhline
\diagbox{Test}{Train} & \dataset & TWH16.2M & Mixamo 
\\
\hline
\dataset            & \bf{1.5 / 0.8} & 2.9 / 1.5 & 2.3 / 1.3 \\
TWH16.2M            & 2.1 / 1.1 & \bf{1.5 / 0.7} & 1.9 / 1.1 \\
Mixamo              & 3.0 / 1.6 & 4.1 / 1.9 & \bf{2.0 / 1.1} \\
\hline\thickhline
\end{tabular}
}
\beforetab
\caption{Cross-dataset evaluation for inside-out body tracking (left) and finger synthesis (right). The prediction is measured by RMSJPE/MPJPE for occluded joints and finger poses in cm, respectively.}
\aftertab
\label{tab:cross_eval}
\end{table*}

\subsection{3-point upper body tracking}

We additionally evaluate our method on 3-point upper body tracking, which is a common missing information scenario for VR headsets that track only the headset and two controllers~\cite{Parger2018,Jiang2016}.
We compare our method against the method of Parger \etal~\cite{Parger2018} on both our \dataset dataset and on Mixamo.
Because the occlusion mask no longer varies between frames, we drop the occlusion mask prediction branch from our method.
In this scenario, we use the transforms of the head-mounted device and the two controllers, since they are tracked by built-in or external sensors. The orientation is represented as forward and upward vectors of the three joints, and used together with their positions as input for the neural network.

As the inverse kinematics approach~\cite{Parger2018} does not support sitting, we also include an evaluation restricted to standing poses. As reported in Table~\ref{tab:3point_pred}, our data-driven method obtains consistent improvement against IK solver-based methods. It is worth noting that the proposed method reduces the reconstruction errors by at least 50\% for the neck, and approximately 66\% for shoulder and elbows.
We further evaluate the generalization ability of our method by performing cross-dataset validation (\ie, training with one dataset and testing on the other). Though the two datasets have quite different statistics, we still observe improvements led by the universal knowledge learned from data and the error is reduced by about 50\% compared with the IK solver-based method (see Table~\ref{tab:3point_pred} \textdagger). Please refer to the accompanying video for detailed comparisons.

\begin{table*}[ptb]
\centering
\setlength{\tabcolsep}{10pt}
\renewcommand\arraystretch{1.15}
\resizebox{0.95\linewidth}{!}{
\begin{tabular}{c|cccc|cccc}
\hline\thickhline
\multirow{2}{*}{Methods} & \multicolumn{4}{c|}{UNOC}  & \multicolumn{4}{c}{MIXAMO}  \\
\cline{2-9}
& Neck & Shoulders & Elbows & All & Neck & Shoulders & Elbows & All \\
\hline
Ours
& \bf{1.1 / 0.9} & \bf{2.7 / 2.3} & \bf{3.5 / 2.9} & \bf{2.3 / 1.7} 
& \bf{1.7 / 1.4} & \bf{4.8 / 4.0} & \bf{6.1 / 4.9} & \bf{4.0 / 2.7} \\
Ours \textdagger
& 1.7 / 1.5 & 4.9 / 1.5 & 6.4 / 5.7 & 4.1 / 3.0 
& 2.7 / 2.3 & 6.7 / 5.8 & 9.9 / 8.7 & 6.3 / 4.6  \\
Parger \etal~\cite{Parger2018}
& 3.9 / 3.5     & 7.5 / 6.2     & 10.3 / 7.8    & 8.3 / 6.3     
& 4.1 / 3.7     & 13.7 / 11.3   & 16.0 / 12.7   & 13.4 / 10.4 \\
Parger \etal~\cite{Parger2018} *
& 2.5 / 2.4     & 7.1 / 5.7     & 7.8 / 5.9     & 6.7 / 5.1 
& 3.3 / 3.0     & 12.2 / 10.0   & 13.9 / 11.4   & 11.8 / 9.2 \\
\hline\thickhline
\end{tabular}}
\beforetab
\caption{Quantitative comparisons of RMSJPE/MPJPE in cm on \dataset for 3 point upper body tracking. Line marked \textdagger~is trained on the other dataset, tested on this dataset. * indicates using standing poses only.}
\aftertab
\label{tab:3point_pred}
\end{table*}


\begin{figure}[ptb]
\centering
\subfloat{\includegraphics[width=0.325\linewidth]{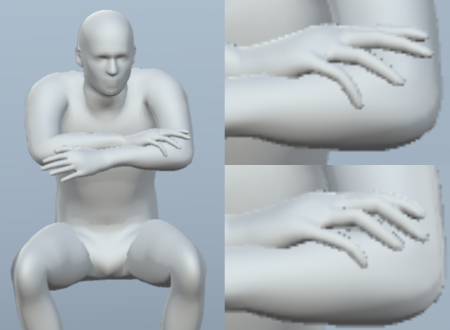}}
\hfill
\subfloat{\includegraphics[width=0.325\linewidth]{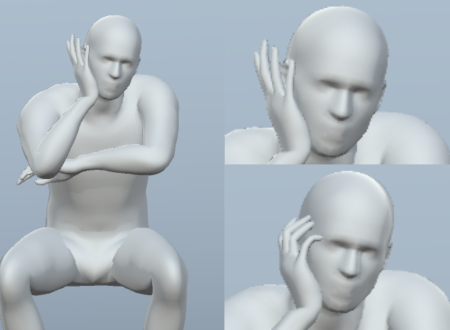}}
\hfill
\subfloat{\includegraphics[width=0.325\linewidth]{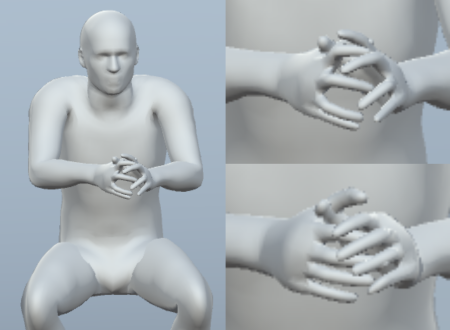}}
\vspace{-8.5pt}
\subfloat{\includegraphics[width=0.325\linewidth]{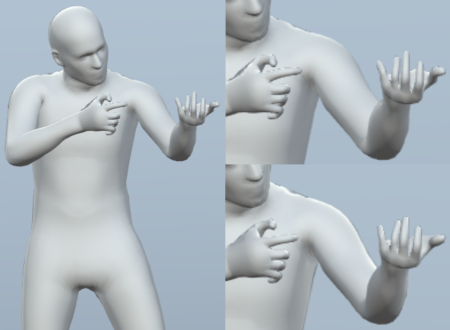}}
\hfill
\subfloat{\includegraphics[width=0.325\linewidth]{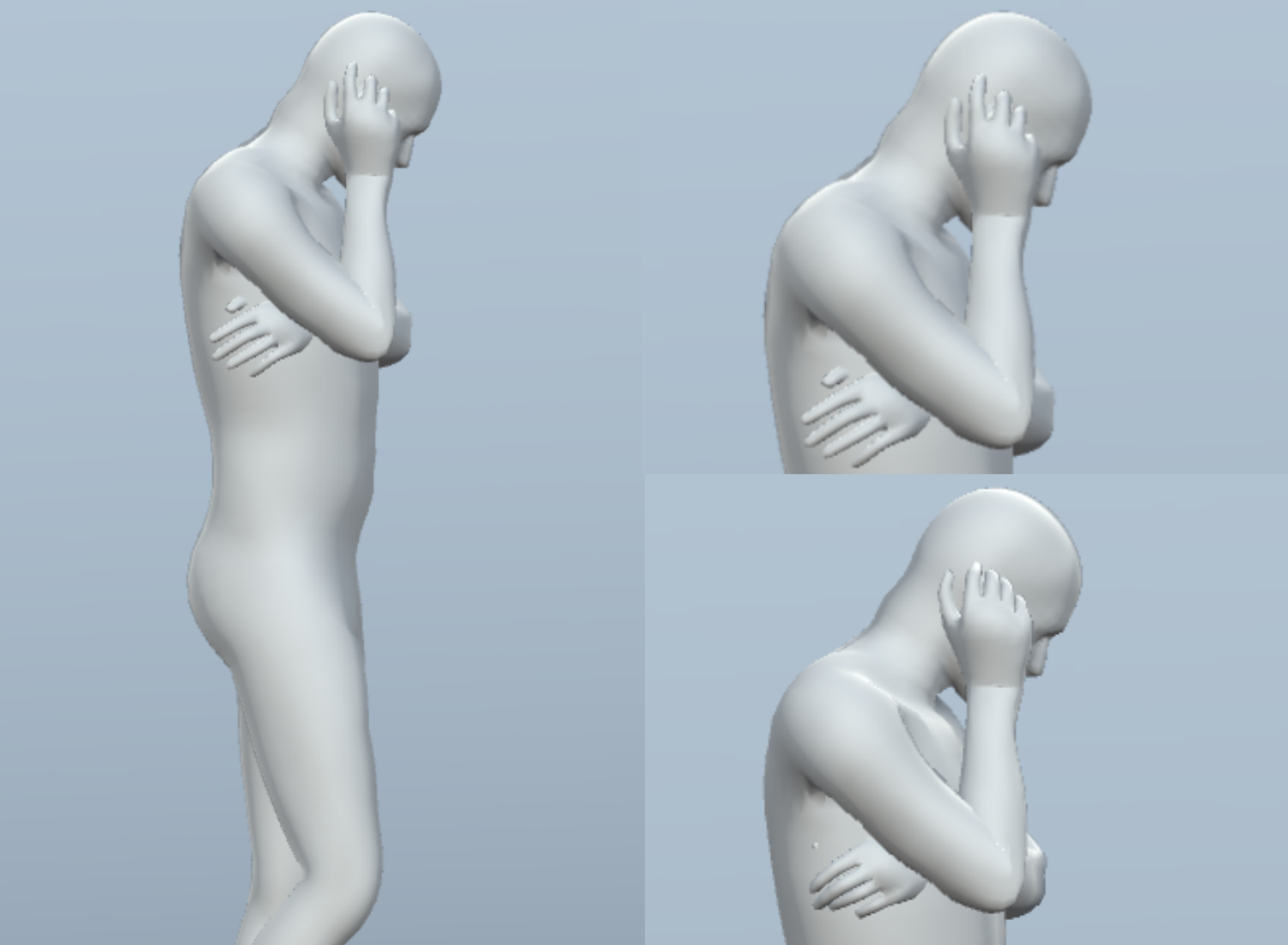}}
\hfill
\subfloat{\includegraphics[width=0.325\linewidth]{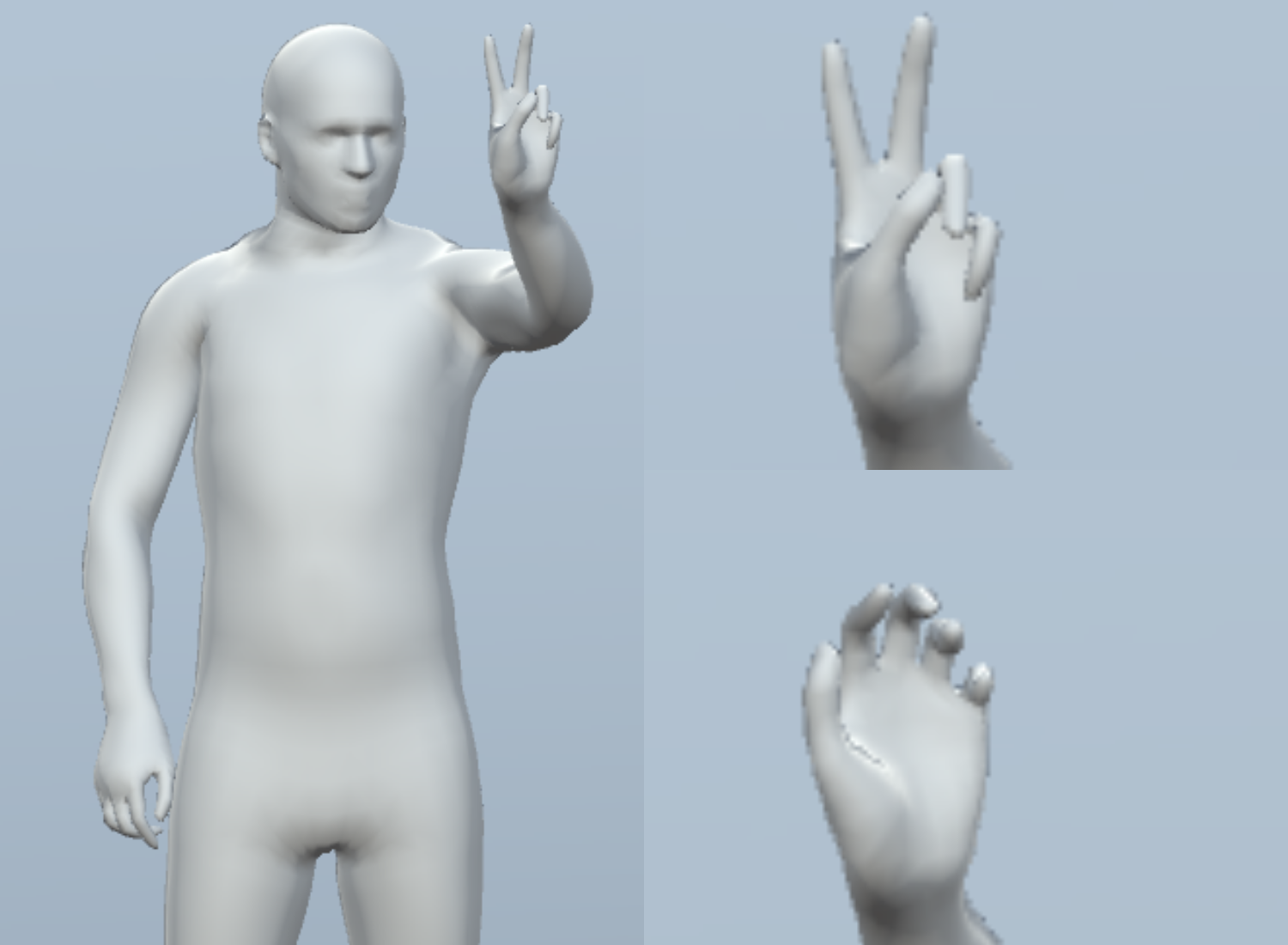}}
\beforefigcaption
\caption{Examples of synthesized finger poses using our model. The avatars on the left show the ground truth, the top right images show a close-up of the ground truth finger pose in comparison to the predicted fingers on the bottom right.}
\afterfigcaption
\label{fig:finger_pred}
\end{figure}

\subsection{Finger synthesis from body motion}

Researchers have proposed various approaches for generating finger motion from body motion, including using database-retrieval methods~\cite{Jorg2012} and audio~\cite{Lee2019}.
However, directly comparing with these approaches is difficult as motion segments, models, code and datasets are seldom available. We slightly adjust our model to solve this task. Specifically, the input for our method is the observed body-only poses (\ie, the positions of all body joints together with the up and forward vectors of the wrists) in a time window of 27 frames, which includes the current frame plus 26 frames of history; 2) the output from our method is the synthesized 32 wrist-local joint positions of the fingers for the current frame. We further include one additional output representing the locator on each finger tip to infer the orientation of the distal bone, resulting in 42 output positions all together.

As can be seen in Fig.~\ref{fig:finger_pred} and in the accompanying video, our network is able to create highly plausible predictions.
This is especially surprising as one cannot assume that finger locations can in general be inferred from general body motions.
We assume that the high quality of the results can be explained by the fact that body motions are often related to achieving a specific goal which includes finger and hand motion.
For example, folding arms, scratching ones head, or leaning on ones hand.
Especially in these hand-body contact situations, our approach achieves highly natural results although no finger information is ever presented during testing.
It is worth noting that our model cannot generate finger animations when the person keeps the static body pose for a while, which is similar to the inside-out body tracking scenario. For example, some hand-only motions, \eg, typing on a keyboard, or scratching the head cannot be synthesized as the model lacks clear semantics about what type of action/interaction is ongoing.

We further show the value of our proposed dataset by cross validating our model across Mixamo, TWH16.2M and our \dataset datasets (see Table~\ref{tab:cross_eval}, right). From the results, we can observe that the model learned on our dataset obtains better generalization abilities than models trained on TWH16.2M and Mixamo datasets. This suggests our dataset covers a unique set of finger motions that has not been captued and covered by other datasets.

\section{Conclusion}

In this paper, we build a motion-capture dataset with simultaneous body and hand tracking.
We demonstrate that our \dataset dataset fills the gap of difficult hand-body interactions and occlusions for the purposes of capturing and predicting rich social interactions.
We show one application of \dataset by training networks to achieve accurate prediction of missing joints. We hope our dataset will promote advancement on body tracking and immersive virtual communication.
The dataset (solved pose and cleaned marker positions), source code and pre-trained models are available at \emph{link removed for review}.


\bibliographystyle{abbrv-doi}

\bibliography{references}
\end{document}